\documentclass[review, 3p, number,sort&compress,11pt]{elsarticle}

\usepackage{hyperref}
\journal{Information Sciences}

\usepackage{upgreek}
\usepackage{amsmath,amssymb,amsfonts}
\usepackage{algorithmic}
\usepackage{graphicx}
\usepackage{subcaption}
\usepackage{physics}
\usepackage{textcomp}
\usepackage{amsthm}
\usepackage{mathtools}
\usepackage[linesnumbered,boxed,lined,,commentsnumbered,ruled,vlined]{algorithm2e}
\usepackage[dvipsnames]{xcolor}
\usepackage{url}
\usepackage{float}
\usepackage{tabularx}
\usepackage{multirow}
\usepackage[normalem]{ulem}
\newtheorem{theorem}{Theorem}
\newcommand*\diff{\mathop{}\!\mathrm{d}}

\usepackage{appendix}
\usepackage{lscape}

\newcounter{ToDo}
\newcounter{gaocomm}
\newcounter{wangcomm}
\newcounter{Note1}
\definecolor{blue-violet}{rgb}{0.54, 0.17, 0.89}
\definecolor{mygreen}{rgb}{0.0, 0.5, 0.0}
\definecolor{awesome}{rgb}{1.0, 0.13, 0.32}
\definecolor{bostonuniversityred}{rgb}{1.0, 0.0, 0.0}

\begin{document}
\begin{frontmatter}

\title{Neural Ordinary Differential Equation Model for Evolutionary Subspace Clustering and Its Applications}


\author[mymainaddress]{Mingyuan Bai\corref{mycorrespondingauthor}}
\cortext[mycorrespondingauthor]{Corresponding author}
\ead{mingyuan.bai@sydney.edu.au}

\author[mymainaddress]{S.T. Boris Choy}
\ead{boris.choy@sydney.edu.au}

\author[mysecondaryaddress]{Junping Zhang}
\ead{jpzhang@fudan.edu.cn}

\author[mymainaddress]{Junbin Gao}
\ead{junbin.gao@sydney.edu.au}

\address[mymainaddress]{Discipline of Business Analytics, The University of Sydney Business School, \\
The University of Sydney, NSW 2006, Australia}

\address[mysecondaryaddress]{Shanghai Key Lab of Intelligent Information Processing and the School of Computer Science,\\
Fudan University, Shanghai 200433, China}

\begin{abstract}
The neural ordinary differential equation (neural ODE) model has attracted increasing attention in time series analysis for its capability to process irregular time steps, i.e., data are not observed over equally-spaced time intervals. In multi-dimensional time series analysis, a task is to conduct evolutionary subspace clustering, aiming at clustering temporal data according to their evolving low-dimensional subspace structures. Many existing methods can only process time series with regular time steps while time series are unevenly sampled in many situations such as missing data. In this paper,  we propose a neural ODE model for evolutionary subspace clustering to overcome this limitation and a new objective function with subspace self-expressiveness constraint is introduced. 
We demonstrate that this method can not only interpolate data at any time step for the evolutionary subspace clustering task, but also achieve higher accuracy than other state-of-the-art evolutionary subspace clustering methods. Both synthetic and real-world data are used to illustrate the efficacy of our proposed method.
\end{abstract}


\end{frontmatter}


\section{Introduction}
Subspace clustering has been an important topic of unsupervised learning over the last two decades. For instance, it plays a crucial role in a number of computer vision tasks such as motion segmentation and facial recognition \citep{Vidal2011}. It also widely used in image representation and compression \citep{Elhamifar09SparseSC}, hybrid system identification \citep{Li2018SystemTheory}, robust principal component analysis \citep{wright2008}, robust subspace recovery and tracking \citep{Li2016RobustRecovery}, recommender system \citep{Agarwal2005}, geographical information systems \citep{Jones2019} and finance \citep{Sim2014}. 

It is common that data are collected in the high-dimensional ambient space while the data themselves have their own intrinsic structures such as being confined in lower-dimensional subspaces. As all data points lie on a union of several subspaces, the core of subspace clustering gathers the data points according to the subspaces that they belong to. That is, the objective of subspace clustering 
is to extract meaningful patterns which have parsimonious and high-dimensional structures. Specifically, subspace clustering assumes that a data point can be described as a linear combination of all other data points in the same subspace. This characteristic is known as the self-expressiveness.

Subspace clustering algorithms are classified into  four major groups: (1) algebraic; (2) iterative; (3) statistical and (4) spectral clustering-based algorithms. As the first three groups of methods have strong model assumptions and suffer from a high computational cost, 
spectral clustering methods have become the most popular subspace clustering algorithms. 
Note that all these algorithms are in principle designed for static data. While for time series data, a straightforward strategy is to apply an individual subspace clustering algorithm on the data at each time step. However, it is evident that this strategy simply ignores the most important evolving information hidden in the time-dependent data. Naturally, it is of great interest 
to develop evolutionary subspace clustering algorithms for time series data. The sparse subspace clustering algorithm \citep{Elhamifar09SparseSC} is a good example.

To understand the process of data evolution and their  specific parsimonious structures, online subspace clustering methods which are built on the spectral subspace clustering algorithms are proposed. A famous online subspace clustering algorithm is the convex evolutionary self-expressive model (CESM) \citep{Hashemi2018EvoSSC} which uses a linearly weighted average smoothing function to obtain the current affinity matrix based on what is in the previous time step. However,
this linear information evolving process cannot be applied to 
non-linear evolving data sequences. A remedy is provided in \citep{Xu2021} where the evolutionary subspace clustering algorithm with the long short-term memory (LSTM-ESCM) model is proposed to efficiently and effectively capture the non-linear sequential parsimonious structure of the time series data.

While we have seen the great success of the new evolutionary subspace clustering methods for time series, we also note that all existing evolutionary subspace clustering methods assume that the time steps are equally-spaced, i.e., $\Delta_m = t_{m+1} - t_{m}$ is a constant. 
In practice, data may not be acquired at some time points and this situation violates the assumption of equally-spaced time steps. Fortunately, neural ordinary differential equations (neural ODEs) \citep{Chen2018NeuralODE} provide a possible solution. A neural ODE models a time-evolving function by learning its derivative (or the vector field) which is a neural network. In the whole time domain, neural ODE can produce the entire evolving processing determined by the initial value, and this is the solution to an initial value problem (IVP).
With this approach, the complicated non-linear sequential features in the data are efficiently and effectively captured, regardless of the irregularity of the time steps. For theoretical properties of neural ODE, as it defines the derivative of the time-evolving function as a neural network which is a universal function approximator, neural ODE can easily provide a consistent solution \citep{Suli2003}. Furthermore, choosing Lipschitz continuous activation functions, such as ReLU, sigmoid and \textit{tanh} if proper scaled, can ensure the unique solution and convergence of neural ODE \citep{Suli2003}.

In this paper, we propose the neural ODE evolutionary subspace clustering method (NODE-ESCM) to address the aforementioned challenges from the equally-spaced time steps assumption and the capability to capture complicated non-linear sequential parsimonious structure. The novelty of this paper is summarized below. 
\begin{enumerate}
    \item  The proposed method inherits the problem formulation of the sparse subspace clustering, designed for both regularly and irregularly sampled time series. Our experiments demonstrate that for time series with irregular time steps, this new method can also achieve the same capability of feature extraction as the case with equally-spaced time steps.
    \item  The affinity matrix in data self-expressiveness is modelled by a well-defined 
    neural ODE. It can be obtained by solving the learned neural ODE 
    without any explicit algorithms, such as the classic subspace clustering algorithms. This learned neural ODE can also provide a unique and consistent solution and converges to the true solution of the IVP.
    \item We propose to use the self-expressiveness constraints in the objective function of the optimization problem. 
    \item In terms of applications, we apply our proposed method to the performance of female's welfare in 19 G20 countries. To the best of our knowledge, this is the first study of the international female's welfare and gender equality using subspace clustering based on deep learning. 
\end{enumerate}

The rest of the paper is organized as follows. Section~\ref{Sec:2} introduces notations and reviews  subspace clustering methods, evolutionary clustering methods and evolutionary subspace clustering methods.
Section~\ref{Sec:3} presents the proposed NODE-ESCM. Section~\ref{Sec:4} demonstrates the flexibility of this model on  experimental design. Synthetic data and real-world data from real-time motion segmentation, ocean water mass study and gender study on female's welfare are analysed using the model. Section~\ref{Sec:5} concludes the paper.

\section{Background}\label{Sec:2}
\subsection{Notation}
We use the boldface capital letters to denote matrices such as $\mathbf{X}$ and the boldface lowercase letters such as $\mathbf{x}$ to represent vectors. $\mathbf{I}$ is the identity matrix of appropriate dimension. A lowercase such as $x$ represents a scalar. Time series and subspaces are denoted by the calligraphic letters such as $\mathcal{X}$ for a sequence of data and $\mathcal{S}$ for a subspace. In general, the dimension of a subspace $\mathcal{S}$ is usually denoted by $d$. 
Consider a time series $\mathcal{X} = \{\mathbf{X}_t\}_{t=1}^T$ at  $T$ time steps where $\mathbf{X}_t$ ($1\leq t\leq T$) 
can be scalars, vectors, matrices or multi-dimensional data, depending on the application scenarios. In this paper, $\mathbf{X}_t$ is a matrix,  consisting of all $D$-dimensional features of $N$ objects 
which belong to the union of $n_t$ 
subspaces. In most cases, $n_t$ is assumed to be a fixed constant. 
We use $\|\quad\|$ to denote a norm, where  $\|\quad\|_1$ is the $\ell_1$-norm and $\|\quad\|_F$ is the Frobenius norm. $\text{tr}(\mathbf{X})$ stands for the trace of matrix $\mathbf{X}$, $\text{diag}(\mathbf{X})$ is  the 
diagonal matrix of $\mathbf{X}$ and  $\text{orth}(\mathbf{X})$ represents the orthogonalization for matrix $\mathbf{X}$, $\text{sign}(x)$ returns the sign of its argument $x$ and $\left \lceil{x}\right \rceil$ gives the integer rounding up to its argument $x$.

\subsection{Subspace Clustering}
Subspace clustering has been used 
in image processing, such as image representation and compression \citep{wright2008,hong2006}. It is also popular in computer vision, such as segmentation of images, motions and videos \citep{Costeira1998SCApp1,Kanatani2001SCApp2,Zelnik-Manor03SCApp3,Yan2006SCApp4,Vidal2008SCApp5}. The subspace clustering problem aims to arrange data into clusters according to their subspace structures, so that different clusters of data belong to different subspaces. Specifically, for data in matrix form $\mathbf{X} = [\mathbf{x}_1, \mathbf{x}_2, ..., \mathbf{x}_N] \in \mathbb{R}^{D\times N}$ where $\mathbf{x}_i\in\mathbb{R}^D$, $i=1,2,\cdots,N$, this problem is to find the number $n\geq 1$ of linear subspaces $\{\mathcal{S}_j\}_{j=1}^n$, $j=1,2,\cdots,n$, where $\mathbf{X}$ are drawn from, their dimensions $\{d_j = \text{dim}(\mathcal{S}_j)\}_{j=1}^n$,
subspace bases $\{\mathbf{U}_j\}_{j=1}^n$, and the segmentation of data points w.r.t. subspaces $\{\mathcal{S}_j\}_{j=1}^n$ \citep{Vidal2011}. Note that the (affine) subspaces are assumed to be  $\mathcal{S}_j = \{\mathbf{x}\in\mathbb{R}^D: \mathbf{x} = \boldsymbol{\mu}_j+\mathbf{U}_j\mathbf{y}\}$ where $\boldsymbol{\mu}_j$ is an arbitrary point in this subspace and it can be set to 0, i.e., $\boldsymbol{\mu}_j = \mathbf{0}$, for linear subspaces, and $\mathbf{y}\in\mathbb{R}^{d_j}$ is a low-dimensional representation of $\mathbf{x}$.

From \citep{Vidal2011}, subspace clustering problems are solved by methods which can be categorized into four taxonomies: (i) algebraic, (ii) iterative, (iii) statistical and (iv) spectral clustering-based methods. Firstly, algebraic methods are mostly designed for linear subspaces and they are not robust to noise \citep{Costeira1998SCApp1,Kanatani2001SCApp2,Boult91,Gear1998,Ichimura1999,Vidal2005GPCA,Ma2008GPCA}. 
For large $d_j$'s and $n$, these methods are computationally expensive or infeasible. 
Secondly, iterative methods are refined to address the sensitivity to noise \citep{Bradley00k-planeclustering,Agarwal2004} .
These methods are sensitive to the initialization and extreme values, and the convergence to a global optimum is not warranted.
Thirdly, statistical methods specify an appropriare generative model for data and thus optimal estimates can be found  \citep{Tipping1999MPPCA,Ma2007ALC,Fischler1981RANSAC}. However, these methods may not be preferred if their distributional assumptions are too strong or too strict such that the optimality of the learning process is unachievable or if the number of the subspaces is large or the dimension of the subspaces is unknown or unequal.

Finally and importantly, spectral clustering-based methods are prominent in high-dimensional data clustering \citep{Elhamifar09SparseSC,Yan06LSA,Zhang2012SLBF,Goh07LLMC}. These methods are easy to implement and are robust to noise and extreme values, and their complexity does not increase with $n$ or $\{d_j\}_{j=1}^n$.
In particular, the sparse subspace clustering (SSC) \citep{Elhamifar09SparseSC} is a  popular subspace clustering algorithm that assumes self-expressiveness and hence each data point in a union of subspaces can be represented by a combination of other data points. That is,
\begin{align}
    \mathbf{X} = \mathbf{X}\mathbf{C},\quad \text{diag}(\mathbf{C}) = \mathbf{0},\label{SSC:1}
\end{align}
where
$\mathbf{C} = [\mathbf{c}_1,\cdots,\mathbf{c}_N]\in\mathbb{R}^{N\times N}$ is the affinity matrix, whose elements  measure the affinity of any two data points in $\mathbf{X}$. For a noisy dataset $\mathbf{X}$, there is no guarantee that $\mathbf{X} = \mathbf{X}\mathbf{C}$.
When the data are distributed according to certain subspace structures,
the affinity matrix $\mathbf{C}$ will have certain block-diagonal structures.
Therefore, the self-expressiveness condition as in Equation~\eqref{SSC:1} can be converted into the following optimization problem. 
\begin{align}
\begin{aligned}
    \min \quad \|\mathbf{C}\|_1+\lambda\|\mathbf{X} - \mathbf{X}\mathbf{C}\|_F^2,\quad
    \text{s.t.}\quad \text{diag}(\mathbf{C}) = \mathbf{0}
\end{aligned}\label{SSC:3}
\end{align}
where $\|\mathbf{C}\|_1$ induces a sparse affinity matrix to simulate the block-diagonal structures. This problem can be solved using an optimization algorithm based on basis pursuit (BP) which is a sparse reconstruction algorithm.
After obtaining $\mathbf{C}$, a symmetric matrix $\mathbf{A}$ is computed from $\mathbf{A} = |\mathbf{C}| + |\mathbf{C}|^\top$ where $|\quad|$ takes the absolute value for each element in $\mathbf{C}$. Then spectral clustering can be applied on $\mathbf{A}$ to cluster data.

To find the affinity matrix $\mathbf{C}$, many methods are proposed to address issues such as the robustness to noise \citep{Liu2010,Liu2013LRSC,Lu13LSR}, avoiding overfitting \citep{Lu13LSR}, capturing relational or structural information \citep{Feng2014RSSBDP,Gao2015MultiviewSSC}, resolving non-convexity \citep{Nie2016SCNLR} and controlling correlations among data points using a threshold \citep{Heckel2015RSCThreshold}. 
Among these methods, the SSC and low-rank representation (LRR) \citep{Liu2010} have been widely generalized in recent years. For example, \citep{Li2017StructuredSSC} enables the representation of learning and clustering to be jointly achieved in SSC-based methods while 
\citep{Elhamifar2016HighRankSC,Li2016StructuredSSC,FAN2017MissSSC,Amin2018SSCMissCorrupt,Tsakiris2018TheoreMissSSC} enhance the capability of handling missing values. A temporal subspace clustering scheme is proposed in  \citep{Li2015TempSSC}. It samples one data point at each time step and aims to assemble data points into sequential segments. Then the segments are clustered into their corresponding subspaces. However, none of these methods considers the sequential relationship based on the evolutionary structure in the feature space. Hence,  evolutionary clustering methods are introduced, aiming at the ability to process the evolutionary structure in the data.

\subsection{Evolutionary Clustering}
Evolutionary clustering is a remarkable technique to process the sequential information in data and incorporate the evolutionary structure. It considers the long-term pattern and is robust to short-term variations. A general framework for the evolutionary clustering is first proposed in \citep{Chakrabarti2006EvoClust}. 
The total cost function, $L_{total}$, in this framework is modified by \citep{Chi2009ModifEvoClust} to 
\begin{align*}
    L_{total} = \alpha L_{temporal} + (1-\alpha) L_{snapshot}
\end{align*}
where $\alpha$ is a weight, $L_{snapshot}$ is the static spectral clustering cost and $L_{temporal}$ has different meanings in their two proposed methods: the Preserving Cluster Quality (PCQ) method and the Preserving Cluster Membership (PCM) method. In the PCQ method, $L_{temporal}$ penalizes the clustering results which do not align with past similarities, while in the PCM method, it penalizes the clustering results which do not align with past clustering results. Here, the total cost is a quadratic objective function.

An intuitive method for time series clustering to find the union of subspaces at different time steps is the concatenate-and-cluster (C\&C) method \citep{Hashemi2018EvoSSC}.
As the computational complexity increases with time, the performance of this method on  capturing the temporal evolution of subspaces is not guaranteed. Therefore, \citep{Xu2014AFFECT} proposes the adaptive forgetting factor for the evolutionary clustering and tracking (AFFECT). The AFFECT constructs an affinity matrix $\mathbf{W}_t$ for each time step $t$. Each element of $\mathbf{W}_t$ represents the similarity between two columns $\mathbf{x}_i$ and $\mathbf{x}_j$ in $\mathbf{X}_t$. This is assumed in the AFFECT that  the affinity matrix can be decomposed as
\begin{align*}
    \mathbf{W}_t = \boldsymbol{\Psi}_t+\mathbf{N}_t,\quad t = 0, 1, 2,\cdots, T,\quad T\in\mathbb{N}
\end{align*}
where the true affinity matrix, $\boldsymbol{\Psi}_t$, is an unknown deterministic matrix of unobserved states and  $\mathbf{N}_t$ is random noise with a zero mean. $\boldsymbol{\Psi}_t$ is estimated by the smoothed affinity matrix $\hat{\boldsymbol{\Psi}}_t$ which is given by 
\begin{align*}
    \hat{\boldsymbol{\Psi}}_t = \alpha_t\hat{\boldsymbol{\Psi}}_{t-1} + (1-\alpha_t)\mathbf{W}_t
\end{align*}
for $t\geq 1$ and $\hat{\boldsymbol{\Psi}}_0 = \mathbf{W}_0$. $\alpha_t$ is a smoothing parameter which is  also known as the forgetting factor at time $t$. $\mathbf{W}_t$ is computed using the information from $\mathbf{X}_t$ by the negative Euclidean distance between its columns or its exponential variant. Note that  past information in the data is not taken into account and therefore, the accuracy of clustering can be jeopardized. Furthermore, the AFFECT determines $\alpha_t$ using an iterative shrinkage estimation approach. This approach is under the strong assumption that $\boldsymbol{\Psi}_t$ has a block structure. This assumption holds only when $\mathbf{X}_t$ is a realization of a dynamic Gaussian mixture model. This is not true in many real-world scenarios for subspace clustering. Prior to the CESM  proposed by \citep{Hashemi2018EvoSSC}, few methods address the estimation of $\alpha_t$.

\subsection{Evolutionary Subspace Clustering }\label{Sec:2:4}
The CESM does not require $\boldsymbol{\Psi}_t$ to have a block structure. It exploits the self-expressiveness property and provides a novel alternating minimization algorithm to estimate  $\alpha_t$. As its name suggests, CESM allows the self-expressiveness property of the data point at each time step, i.e.,
\begin{align}
    \mathbf{X}_t = \mathbf{X}_t\mathbf{C}_t,\quad \text{diag}(\mathbf{C}_t) = 0 \label{TempSSC}
\end{align}
where $\mathbf{X}_t = [\mathbf{x}_{t,1}, \mathbf{x}_{t,2},\cdots,\mathbf{x}_{t,N_t}]\in\mathbb{R}^{D_t\times N_t},\quad t = 0, 1, 2,\cdots, T$. Static clustering algorithms can be used to estimate $\mathbf{C}_t$. However, they fail to capture the evolutionary structure of the time series data as only information from the current time step is taken into account. Therefore, evolutionary subspace clustering methods are proposed. For example, see \citep{Hashemi2018EvoSSC}. Under these methods,   
\begin{align}
\begin{aligned}
    &\mathbf{C}_t = f_\theta(\mathbf{C}_{t-1}),\quad\mathbf{X}_t = \mathbf{X}_t\mathbf{C}_{t},\\
    &\quad\text{diag}(\mathbf{C}_t)=0,\quad t=0, 1, 2,\cdots, T,\quad T\in\mathbb{N}.
\end{aligned}    \label{CESM:basic}
\end{align}
where $f_\theta:\mathcal{P}_\mathcal{C}\rightarrow\mathcal{P}_\mathcal{C}$ is a matrix-valued function parameterized by $\theta$ and it processes the self-expressiveness property. The domain and the range of $f_\theta$ are  $\mathcal{P}_\mathcal{C}\subset\mathbb{R}^{N\times N}$ which includes  any chosen parsimonious structures imposed on the representation matrices at each time point, such as  sparse or low-rank representations. Thus, the optimization problem is formulated as 
\begin{align}
    \min_\theta &\quad \|\mathbf{X}_t - \mathbf{X}_tf_\theta(\mathbf{C}_{t-1})\|_F^2\notag\\
    \text{s.t.}&\quad f_\theta(\mathbf{C}_{t-1})\in\mathcal{P}_\mathbf{C},\quad \mathbf{C}_{t} = f_\theta(\mathbf{C}_{t-1}).
\label{CESM:optima}
\end{align}
Spectral clustering is applied to  the affinity matrix $\mathbf{A}_t = |\mathbf{C}_t|+|\mathbf{C}_t|^\top$ and  the union of the subspaces can then be obtained.
Specifically, \citep{Hashemi2018EvoSSC} proposes the following convex function for $f_\theta$:   
\begin{align*}
    \mathbf{C}_t = f_\theta(\mathbf{C}_{t-1}) = \alpha\mathbf{U}+(1-\alpha)\mathbf{C}_{t-1}
\end{align*}
where $\theta = (\mathbf{U},\alpha)$ and $\mathbf{U}$ is the innovation representation matrix to capture the drifts between two consecutive time points. Alternative minimization algorithms are proposed for this evolutionary subspace clustering problem. Although this evolutionary subspace clustering method does not require any distributional assumptions for the data, the estimation of the current representation matrix relies on the immediately past information from the evolving data and a linear function $f_\theta$. Hence the nonlinear information and long memory property in the evolving data are neglected. Consequently, the representation of self-expressiveness is affected and  therefore,  a nonlinear evolutionary subspace clustering algorithm with long memory property is in demand.

\subsection{LSTM Evolutionary Subspace Clustering}
Neural networks are a universal function approximator. They can process complicated  nonlinear information hidden in the data. In the neural networks family, long short-term memory (LSTM) model is a recurrent neural network and a classical method to capture sequential information from both short-term and long-term dependencies. \citep{Xu2021} proposes long short-term memory evolutionary subspace clustering method (LSTM-ESCM) which incorporates the LSTM model into the evolutionary subspace clustering algorithms. Unlike CESM, LSTM-ESCM defines $f_\theta$ as LSTM networks and the hidden state $\mathbf{C}_t$ as $\mathbf{C}_t = f_\theta(\mathbf{C}_{t-1},\mathbf{X}_t)$. Here, $\theta$ is the set of all parameters in an LSTM and 
$\mathbf{C}_t$ is the self-expressive matrix for LSTM-ESCM. Hence, the proposed sparse optimization regime is 
\begin{align*}
    \min_\theta &\quad \|\mathbf{X}_t-\mathbf{X}_tf_\theta(\mathbf{C}_{t-1},\mathbf{X}_t)\|_F^2+\lambda\|\mathbf{C}_t\|_1\\
    \text{s.t.}&\quad \text{diag}(\mathbf{C}_t)=0,\quad \mathbf{C}_t = f_\theta(\mathbf{C}_{t-1},\mathbf{X}_t).
\end{align*}
This optimization regime not only exploits the nonlinear self-expressiveness from the evolving data but also captures the sequential information. However, the assumption of equally-spaced time steps in LSTM-ESCM makes it less flexible under an irregularly-sampled time series scenario. Such a challenge should be addressed and new techniques need to be developed.

\section{Neural ODE Model for Evolutionary Subspace Clustering}\label{Sec:3}
In this section, we propose the neural ODE evolutionary subspace clustering method (NODE-ESCM). This method can overcome the challenges of irregularly-sampled time series. It also 
identifies sequential data patterns and captures the nonlinear information and evolutionary structure in subspace clustering problems. 

\subsection{Preliminaries}
As aforementioned, neural ODE is a time series method without the equally-spaced time steps assumption. In addition, it also belongs to  a family of deep neural network models which can be regarded as a continuous equivalent to the Residual Networks (ResNets) \citep{Chen2018NeuralODE}. 
Unlike the ResNets, which assumes equally-spaced time steps, neural ODE considers the difference between $t$'s as $\Delta t \rightarrow 0$ such that 
\begin{align*}
    \dv{\mathbf{h}(t)}{t} = g(\mathbf{h}(t),\theta,t),
\end{align*} 
where $\mathbf{h}(t)\in\mathbb{R}^{m}$ and the continuous dynamics of hidden units are parameterized using an ODE specified by the neural network $g(\mathbf{h}(t),\theta,t)$. Note that at each $t$ there are shared parameters $\theta$. We can also write $g(\mathbf{h}(t),\theta,t)$ as $g_\theta(\mathbf{h}(t),t)$.
As neural ODE allows for continuous and infinite number of layers or time steps, for any time $T\in\mathbb{R}$, we can transform a data point $\mathbf{h}_0$ to a set of features $\phi(\mathbf{h}_0)$ by solving an initial value problem (IVP)
\begin{align*}
    \dv{\mathbf{h}(t)}{t} = g_\theta(\mathbf{h}(t),t),\quad \mathbf{h}(0) = \mathbf{h}_0.
\end{align*}
To find the solution to this IVP, \citep{Chen2018NeuralODE} suggests the ODE solver $ODESolve$ to find $\theta$, where a fast backpropagation has been established based on the adjoint sensitivity method.

Evidently, neural ODE is a special ODE and hence inherits the theoretical properties of ODE. Firstly, neural ODE can obtain the unique solution if it satisfies the conditions in Picard's existence theorem \citep{Suli2003}.
\begin{theorem}(Picard's Theorem) Suppose that the real-valued function $(x, y)\rightarrow f(x, y)$ is continuous in the rectangular region $R$ defined by $x_0\leq x \leq X_M$, $y_0-C\leq y \leq y_0 + C$; that $|f(x, y_0)|\leq K$ when $x_0\leq x \leq X_M$; and that $f$ satisfies the Lipschitz condition. There exists $L > 0$ such that $$|f(x, u) - f(x, v)|\leq L|u - v|\quad \text{for all}\quad (x, u)\in R.$$ Assume further that $C\geq\frac{K}{L}(e^{L(X_M-x_0)}-1).$ Then there exists a unique function $y\in C^1 [x_0, X_M]$ such that $y(x_0)=y_0$ and $y' = f(x,y)$ for $x\in[x_0, X_M]$. Moreover, $$|y(x) - y_0|\leq C, \quad x_0\leq x\leq X_M.$$ \label{thm1}
\end{theorem}
For the consistency of neural ODE, the approximation method of its solver, such as the Runge-Kutta methods \citep{Suli2003}, determines the conditions which neural ODE easily satisfies. The convergence requires neural ODE to satisfy the conditions in Picard's theorem and consistency. 

Neural ODE combines the ODE and neural networks in modern machine learning. It  has a remarkable contribution to memory efficiency, adaptive computation and scalable and invertible normalizing flows from its backpropagation. Unlike the ResNet and recurrent neural networks, neural ODE provides practical advantages in continuous time settings. In addition, it also inherits the consistency from ODE, and provides theoretical convergence when the activation functions of $g_{\theta}$ are Lipschitz continuous and $g_{\theta}$ and $\mathbf{h}(t)$ are bounded, required by Picard's theorem and the convergence conditions \citep{Suli2003}.

\subsection{Neural ODE Evolutionary Subspace Clustering Method}
\begin{figure}[tbh]
    \centering
    \includegraphics[height=3cm, width=16.5cm]{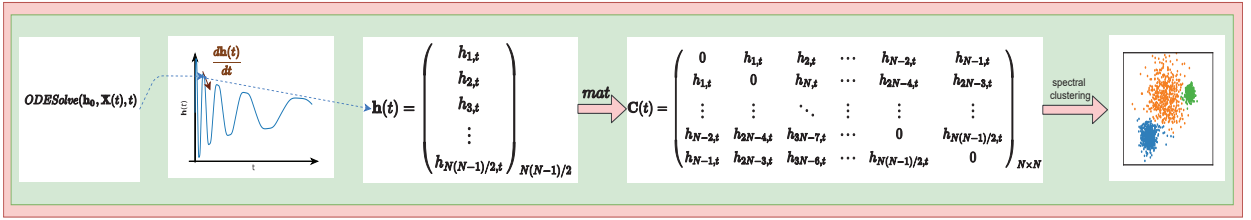}
    \caption{The process to compute the clustering results for each time point $t$, given the ODE and neural ODE solver $ODESolve$ and the learned parameters $\theta$}
    \label{fig:pipeline}
\end{figure}

In Section~\ref{Sec:2:4}, the optimization problem for evolutionary subspace clustering is formulated in Equation~\eqref{CESM:optima}. Assuming equally-spaced time steps, 
$f_\theta$ is either linear \citep{Hashemi2018EvoSSC} or LSTM \citep{Xu2021}. In other words, the data dynamic is  described by a recursive iteration over equally-spaced discretized time steps, $\mathbf{C}_t = f_\theta(\mathbf{C}_{t-1})$.
In the context of neural ODE, we propose a continuous dynamic for $\mathbf{C}(t)$. Instead of estimating each $\mathbf{C}_t$, the estimation of $\mathbf{C}(t)$ is through $\theta$ which is the solution of the neural ODE. This new method is named as the neural ODE evolutionary subspace clustering method (NODE-ESCM). Let the structure set $\mathcal{P}_{\mathbf C}$ be an $N\times N$ symmetric matrix whose diagonal elements are 0, and $\mathbf{h}(t)\in\mathbb{R}^{N(N-1)/2}$ whose elements are stacked from the lower-triangular part of $\mathbf{C}(t)$. 
Then, specifically, we define the dynamic ``curve'' $\mathbf{C}(t)$ through $\mathbf{h}(t)$. In fact,  $\mathbf{h}(t)$ is the solution of the following neural ODE
\begin{align}
    \dv{\mathbf{h}(t)}{t} = g_\theta(\mathbf{h}(t),\mathbf{X}(t),t),\quad \mathbf{h}(0) = \mathbf{h}_0. \label{Eq:16}
\end{align}
where $t\in[0, T]$ and $ T\in\mathbb{R}$. $\mathbf{h}(0) = \mathbf{h}_0$ is the initial value of the neural ODE. $\mathbf{X}(t)$ is the interpolation of input data $\mathbf{X}_t, t=0,1,\cdots,T$, as required by many ODE and neural ODE approximation methods, such as the Runge-Kutta methods, to obtain the numerical solutions. $\mathbf{X}(t)$ can also be regarded as a control variable. The reason to include $\mathbf{X}(t)$ is that we can consider more past information in the time series in order to capture more sequential information. $g_\theta(\mathbf{h}(t), \mathbf{X}(t))$ is a neural network and $\theta$ contains all parameters of the network. This neural network takes the following form
\begin{align}
    g_\theta (\mathbf{h}(t), \mathbf{X}(t), t) = & \underbrace{\sigma\circ\cdots\circ\sigma\circ}_{L-2}\sigma(\mathbf{W}_{2}(\sigma(\mathbf{W}_{11}\mathbf{h}(t))\odot \sigma(\mathbf{W}_{12}\text{vec}(\mathbf{X}(t)))))\label{Eq:17}
\end{align}
where $L$ is the number of layers, $\sigma(\cdot)$ is an activation function, $\theta = \{\mathbf{W}_{11}, \mathbf{W}_{12}, \mathbf{W}_2,\cdots,\mathbf{W}_{L}\}$, $\text{vec}(\cdot)$ is the operator to flatten the matrix $\mathbf{X}(t)$ into a vector and $\odot$ is the Hadamard product, i.e., the element-wise product. Here we choose the Lipschitz continuous activation functions such as ReLU, sigmoid and \textit{tanh} with batch normalization for each layer to ensure the unique solution and convergence as required in Theorem~\ref{thm1}. 
Hence, the solution $\mathbf{h}(t_1)$ is
\begin{align}
    \mathbf{h}(t_1) &= \mathbf{h}(0) +\int_{0}^{t_1} g_\theta(\mathbf{h}(t), \mathbf{X}(t), t)\diff t
\end{align}
where $t_1\in[0, T]$. In our experiments, we choose a linear function for the activation function in the last layer $L$.
In subsequence, 
$\mathbf{h}(t), t \in [0,T],$ can be obtained using $ODESolve$, i.e. 
\begin{align}
    \mathbf{h}(t) = ODESolve(\mathbf{h}_0, \mathbf{X}(t),t) \label{Eq:19}
\end{align}
where 
we can either set $\mathbf{h}_0 = \mathbf 0$ or randomly generate it. For the sake of convenience, we denote the transformation from $\mathbf{h}(t)$ to the symmetric matrix $\mathbf{C}(t)$ by 
\begin{align*}
    \mathbf{C}(t) = \text{mat}(\mathbf{h}(t)).
\end{align*}  
In fact, this transformation is a linear mapping of the solution $\mathbf{h}(t)$ to the structured matrix $\mathbf{C}(t)$. It is not hard to work out $\frac{d \mathbf{C}(t)}{d\mathbf{h}(t)}$\footnote{We use PyTorch to implement this step.}. Given $\mathbf{C}(t)$, we can apply a standard spectral clustering method to it and obtain the subspace clustering results at each time point $t$. For the whole process to compute the clustering results given the learned $\theta$ for $ODESolve$, it is summarized as in Figure~\ref{fig:pipeline}.
With the purpose of learning $\theta$ for $ODESolve$ and obtaining accurate clustering results, we should properly define the neural ODE evolutionary subspace clustering problem.

Hence, for a given set of data sequence $\mathcal{X} = \{\mathbf X_1, \mathbf X_2, ..., \mathbf X_T\}$ observed at discrete time points $\mathcal{T} = \{t_1, t_2, ..., t_T\}$ over the time interval $[0, 1]$, the neural ODE evolutionary subspace clustering problem is 
\begin{align}
\begin{aligned}
    \min_{\theta}&\quad \frac{1}{2} \sum_{j=1}^T(t_{j} - t_{j-1}) \|\mathbf{X}_{t_j}-\mathbf{X}_{t_j}\mathbf{C}(t_j)\|_F^2+\lambda \sum_{j=1}^T\|\mathbf{C}(t_j)\|_1, \\
    \text{s.t.} &\quad  \mathbf{C}(t_j) =  \text{mat}(ODESolve(\mathbf{h}_0, \mathcal{X}, t_j)).
\end{aligned}\label{Eq:20a}    
\end{align}
A regularization term of $\ell_1$-norm is added to the objective function to impose the sparsity of the coefficient $\mathbf{C}(t)$ and avoid the overfitting. $\lambda>0$ is the corresponding regularizer. 

If the set of data sequence is observed at equally-spaced time points $\mathcal{T} = \{0, 1, 2, ..., T\}$, the loss function becomes 
\begin{align}
L(\theta) = \frac1{2(T+1)}\sum^{T}_{t=0} \|\mathbf X_t - \mathbf X_t\mathbf{C}(t)\|^2_F + \lambda \sum^{T}_{t=0}\|\mathbf{C}(t)\|_1. \label{Eq:20}
\end{align}
Hence learning the affinity matrix $\mathbf{C}(t)$ for the evolutionary subspace clustering can be regarded as a deep learning problem whose loss function is given in Equation~\eqref{Eq:20}. Let $L(\theta)= \sum_{t=0}^{T} l_t$ where 
\begin{align*}
    l_t &=
    \frac{1}{2}\text{tr}(\mathbf{X}_t^\top\mathbf{X}_t(\mathbf{I}-\mathbf{C}(t))(\mathbf{I}-\mathbf{C}(t)^\top))+\lambda \|\mathbf{C}(t)\|_1.
\end{align*}
As $\mathbf{X}_t$ is a known constant, the first derivative of $l_t$  w.r.t. $\mathbf{C}(t)$ is given by 
\begin{align*}
    \dv{l_t}{\mathbf{C}(t)}
    =& \mathbf{X}_t^\top\mathbf{X}_t(\mathbf{C}(t)-\mathbf{I})+\lambda \, \text{sign}(\mathbf{C}(t)),   \label{Eq:28}
\end{align*}
Applying the chain rule, the gradient of the loss function is  
\begin{align*}
    \dv{L(\theta)}{\theta}
= \sum_{t=0}^{T} \dv{l_t}{\mathbf{C}(t)}\dv{\mathbf{C}(t)}{\mathbf{h}(t)}\dv{\mathbf{h}(t)}{\theta}. 
\end{align*}
Based on the adjoint method \citep{Chen2018NeuralODE}, the backpropagation algorithm for neural ODE can efficiently evaluate $\dv{\mathbf{h}(t)}{\theta}$.
The gradient descent algorithm is then used to update the estimates of $\theta$ and $\mathbf{C}_t$. 
Specifically, we initialize the parameters at the first iteration using  random generation as in Equation~\eqref{Eq:19} and we use our proposed method to finalize the whole learning process of finding $\mathbf{C}(t)$. The entire learning process is summarized in Algorithm~\ref{Alg:1}.  At each time point, the standard spectral clustering algorithm is applied to $\mathbf{C}(t), t=0,1,\cdots,T$ to obtain the subspace clustering for the data $\mathbf{X}_t$.

\textit{Remark 1:} Our algorithm can deal with observations at irregular time steps if we use the loss defined in Equation~\eqref{Eq:20a}. This will be demonstrated in Section~\ref{Sec:4:3}.

\textit{Remark 2}: For a given set of data, we know the neural ODE defined in Equation~\eqref{Eq:16} after learning the model. With this newly learned neural ODE, we estimate $\mathbf{h}(t)$ at any time point using observation $\mathbf{X}_t$ and a chosen (numerical) ODE and neural ODE solver. Hence, the data affinity $\mathbf{C}(t)$ is obtained and can be applied to subspace clustering at the time $t$. This is not possible for any other  evolutionary subspace clustering strategies. See Section~\ref{Sec:4:3} for details. 

\begin{algorithm}
\SetKwData{Left}{left}\SetKwData{This}{this}\SetKwData{Up}{up}\SetKwFunction{Union}{Union}\SetKwFunction{FindCompress}{FindCompress}\SetKwInOut{Input}{Input}\SetKwInOut{Output}{Output}
\SetAlgoLined
\Input{Training dataset $\mathcal{X}=\{\mathbf{X}_t\}_{t=0}^T$, the number of epochs $itr$, the hidden layer size $h$, the regularization parameter $\lambda$, the learning rate scheduler parameter $\gamma$ and the step size $s$ in $ODESolve$.}
\Output{$\mathbf{C}(t)$ and $\theta$}
\textbf{Initialization}: $\mathbf{h}(0)=\mathbf 0$, and randomly generate $\theta_0$ \\
\For{$i = 1,\cdots,itr$}{
   run the ODESolve with the given step size $s$ in which using Equation~\eqref{Eq:17} to calculate the network $g_{\theta}$.\\
   \noindent run BP for neural ODE to update all network parameters $\theta$, e.g., with ADAM optimizer.
 }
 \caption{Neural ODE Evolutionary Subspace Clustering Algorithm}\label{Alg:1}
\end{algorithm}

\section{Experiments}\label{Sec:4} 
Our proposed NODE-ESCM enables the flexibility of time steps for subspace clustering problems. This method does not require  the assumption of equally-spaced  time steps which is common in LSTM-ESCM and other existing evolutionary subspace clustering methods. To the best of our knowledge, none of the existing methods can handle irregularly-sampled time series.

In this section, our investigation is divided into three parts. In Part 1, we perform a simulation study and assess the performance of NODE-ESCM.
In Part 2, we compare the performance of NODE-ESCM on real-time motion segmentation tasks with other methods. In Part 3, 
 we analyze real data in an ocean water mass study and examine our method in a real-world scenario. It is designed to testify that the NODE-ESCM can provide informative results for all time steps  while known information is given at irregular time steps. Last but not least, we evaluate the performance of our method on female's welfare, which is a highly debatable topic in social science. We investigate whether new information will emerge and whether our findings are consistent with the social science knowledge. 

\subsection{Simulation Study}
\subsubsection{Experimental Setup}
In this simulation study, we consider five $4$-dimensional subspaces  from $\mathbb{R}^{30}$ and 21 synthetic data points are sampled at random from each subspace. For $i = 1,2,...,5$, let $\mathbf{Q}_{i} = \text{randn}(4,21)$ and $\mathbf{U}_i = \text{orth}(\text{randn}(30,4))$ be an orthonormal basis. 
Suppose that the total number of time instances is $10$. At the first time instance,
$t = 1$, the data in the $i^{\rm th}$ subspace is given by 
\[
\widehat{\mathbf{X}}_i = \mathbf{U}_i\mathbf{Q}_i+0.1\times \text{randn}(30, 21),\quad i=1,2,\cdots,5
\]
where $\widehat{\mathbf{X}}_i\in\mathbb{R}^{30\times 21}$. Hence, the data at $t = 1$ is 
\[
\mathbf{X}_{1} = [\widehat{\mathbf{X}}_1, \cdots, \widehat{\mathbf{X}}_5] \in \mathbb{R}^{30\times 105}.
\]
In practice, we can shuffle the columns of $\mathbf X_1$ as the data in each subspace are randomly generated in the same way.    
Using $\mathbf{X}_{1}$, we can then generate an evolving sequence for the remaining 9 time instances. 
The data for the next time instance $t+1$ can be obtained via a rotation of the data at time instance $t$. Here, we use a Givens rotation $\mathbf{G}_t\in\mathbb{R}^{30\times 30}$ with a randomly generated angle $\theta$ from a uniform $U\left(0, \pi/10  \right)$ distribution for each time instance. Hence, the data at time $t+1$ is 
\[
\mathbf{X}_{t+1} =  \mathbf{G}_t\mathbf{X}_t,\quad t = 1, \cdots, 9.
\]
In consequence, we generate a time series
$\mathcal{X} = \{\mathbf{X}_{t}\in \mathbb{R}^{30\times105}\}_{t=1}^{10}$ 
whose elements lie on 5 distinct subspaces.

\subsubsection{Tasks}
This simulation study aims to demonstrate the ability of the NODE-ESCM in handling an evolving dataset, examine the sensitivity of the NODE-ESCM to the sparsity penalty term $\lambda$ and evaluate the accuracy of the NODE-ESCM on ground truth labels. Given the synthetic time series, we conduct clustering for each $\mathbf{X}_t$. First we use the training error $MSE_{\text{training}}$, which is the first term in Equation~\eqref{Eq:20}, for each epoch to measure the convergence. In this study, the number of epochs is set to $100$. To evaluate the model sensitivity to $\lambda$, a $95\%$ confidence interval of the training error is used.
Here, possible values of $\lambda$ are 0, 0.05, 0.10, ..., 2. 
For each $\lambda$ value, we compute the training errors along all epochs and then the mean training error and a $95\%$ confidence interval are reported. For other model hyperparameters, the number of hidden layers is $2$ and each layer has size $ h = 40$. The learning rate scheduler is  $\gamma = 0.1$ and its step size is $s = 100$.
The second evaluation criterion is the clustering accuracy.
For this criterion, we compare the performance of the NODE-ESCM and the LSTM-ESCM on the  synthetic data.

\subsubsection{Simulation Results}
\begin{figure}
  \begin{subfigure}[t]{0.45\columnwidth}
    \includegraphics[trim={2.2cm, 0.2cm, 3.5cm, 1.9cm},clip, width=\linewidth]{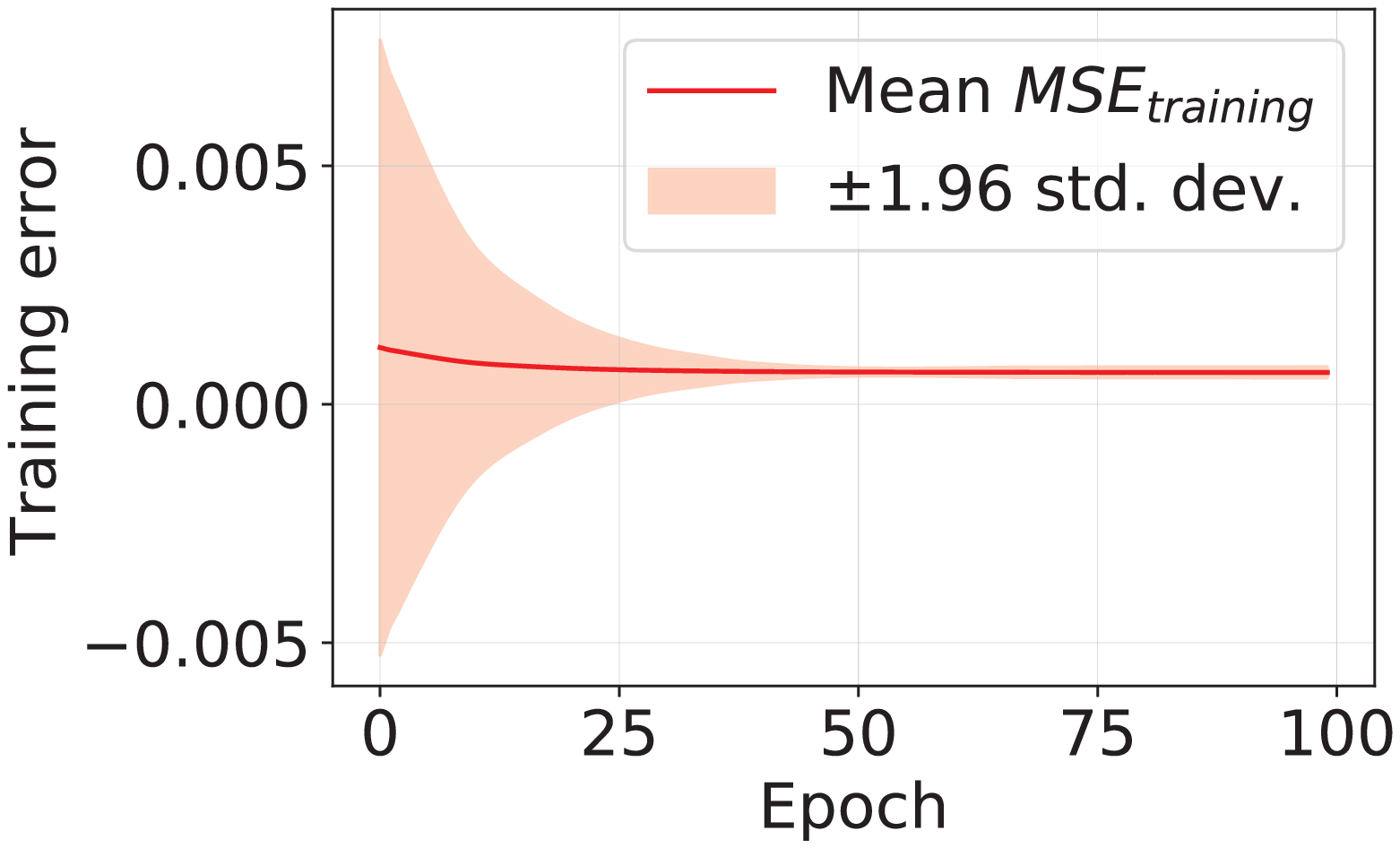}
    \caption{Sensitivity analysis of training errors $MSE_{training}$ for the $40$ $\lambda$'s with the solid line as their mean and the shaded area as the $95\%$ confidence interval
    }\label{Fig:1a}
  \end{subfigure}
  \hfill 
  \begin{subfigure}[t]{0.45\columnwidth}
    \includegraphics[width=\linewidth]{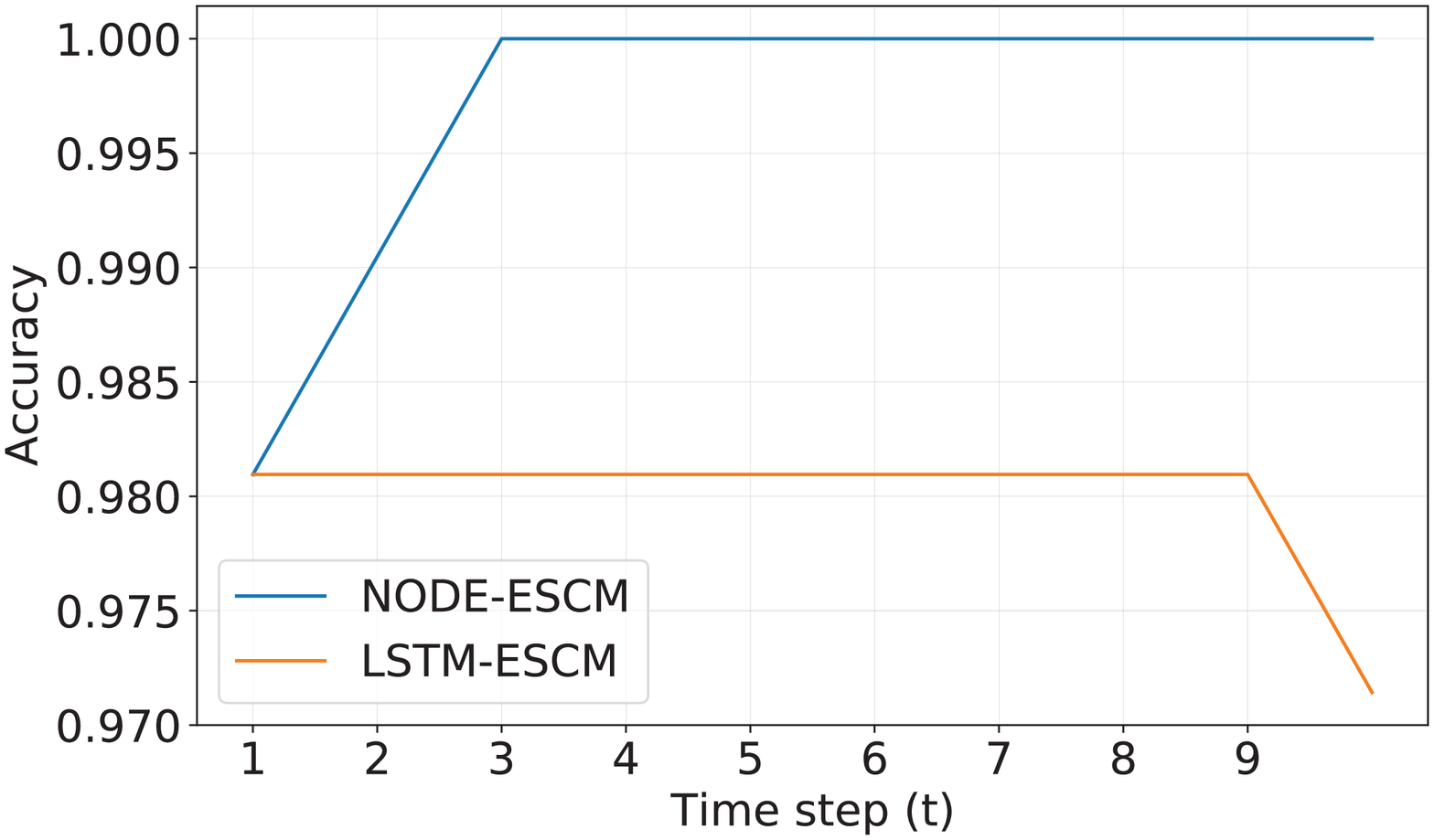}
    \caption{Accuracy convergence along time steps}
    \label{Fig:1b}
  \end{subfigure}
  \caption{Convergence of training error and accuracy in a synthetic data analysis}\label{Fig:1}
\end{figure}

For the first evaluation criterion, the convergence of the training error $MSE_{training}$ is presented in Figure~\ref{Fig:1}(a).
Across these $100$ epochs, it can be observed that the empirical speed of the convergence is evidently fast for the training error. The mean training error also decays smoothly and converges to its optimum at the $20^{\rm th}$ epoch. Furthermore, although the length of the $95\%$ confidence interval is the longest one at 0.0129 for the first epoch, it smoothly decays to 0.0002 after the $50^{\rm th}$ epoch while the mean training error converges to 0.0007. These results also indicate that the NODE-ESCM is generally insensitive to the choice of 
$\lambda$.

For the second evaluation criterion, we compare the clustering accuracy at $t=1,\cdots,10$, for the NODE-ESCM and the LSTM-ESCM at their optimal hyperparameter values after grid search, and the best parameter value $\lambda=1.0$ (NODE-ESCM) and $0.1$ (LSTM-ESCM), respectively.
Figure~\ref{Fig:1}(b) displays the clustering accuracy at different time steps. 
The NODE-ESCM attains a 100\% accuracy from the $3^{rd}$ time step onward while the LSTM-ESCM maintains its accuracy at around 98.1\%. 
It also reveals that the LSTM-ESCM accumulates the previous errors when learning the following time instances, whereas the NODE-ESCM is able to capture the long-term evolving sequential information. Therefore, for long-term learning tasks, the NODE-ESCM is preferred to the LSTM-ESCM. 

To conclude, the proposed NODE-ESCM achieves a fast convergence in the training process and an accurate clustering performance for almost  all time instances, in the simulation study. It is also shown that the NODE-ESCM outperforms the LSTM-ESCM according to the cluster accuracy criterion.

\subsection{Empirical Study 1: Real-Time Motion Segmentation}
\subsubsection{Experimental Setup}
In this empirical study, we consider motion segmentation in videos. From computer vision, we understand that the trajectory vectors of keypoints on the same moving object fall into a lower dimensional subspace of dimension 3 or 4. Before applying the NODE-ESCM here, the given videos are preprocessed as follows.

First, we convert each video (i.e. a stream of frames) into a new time series of length $T$,  i.e. $\mathcal{X}=\{\mathbf{X}_t\}_{t=1}^T$ where  $\mathbf{X}_t$ is the $t^{\rm th}$ data snapshot consisting of $F$ consecutive frames of the video from time $t$. Its dimension is hence $2F\times N$ where $N$ is the number of tracked keypoints and the $(x,y)$-coordinates of the keypoints in a frame are recorded in 2 rows. Therefore, we have $N$ vectors in $\mathbb{R}^{2F}$. The movement segmentation can then be reformulated as a subspace clustering problem given the fact from computer vision that the trajectory vectors in $\mathbb{R}^{2F}$ from the same movement fall into a lower dimensional subspace. For each snapshot $\mathbf{X}_t$ of $F$ consecutive frames, we assume $F \geq 2n$ for a video with $n$ motions.

In this study, we do not simply learn a solver for each particular video sequence. Instead we establish the NODE-ESCM for a number of motion sequences so that the learned NODE can be used as a universal solver for the affinity matrix $\mathbf{C}_t$ of $\mathbf X_t$ for any sequences from similar video contexts. Hence, we train the NODE-ESCM as a whole using the sequence data from multiple motion sequences. To make the training task a bit easier, we assume that all video sequences have the same number of keypoints to be segmented into different motions. In other words, $N_i$ are the same for all $\{\mathbf{X}^i_{t}\}_{t=1}^T$ where $i$ is the index of the given videos\footnote{When we talk about a particular video sequence, we ignore the index $i$.}, and $n_i$ (number of movements in the $i^{\rm th}$ video) are either $2$ or $3$. In the Hopkins 155 dataset, the number of keypoints for each sequence ranges from $63$ to $548$, so we set $300$ keypoints as our target for qualified sequences. To construct a dataset with $N_i = 300$, we first select $21$ specific sequences with $N_i$  between $300$ and $350$ keypoints. Table~\ref{Tab:Hopkins155} presents the 
the average number of keypoints and the average number of frames of these $21$ selected sequences with either $2$ or $3$ motions. Afterwards,  we set a limit of $300$ keypoints and randomly eliminate the same number of excess keypoints from each sequence. Then, we select $17$ sequences at random as the training set and treat the remaining $4$ sequences as the test set. Finally, the NODE-ESCM solver is trained using the whole training set altogether.
\begin{table}
\centering
\caption{Distribution of the number of points and frames.}
\label{Tab:Hopkins155}
\begin{tabular}{c|ccc|ccc}
\hline
             & \multicolumn{3}{c|}{2 Groups} & \multicolumn{3}{c}{3 Groups}  \\
             & \# Seq. & Keypoints & Frames     & \# Seq. & Keypoints & Frames     \\
\hline
Checkerboard       & 14      & 333    & 28         & 4      & 320    & 31         \\
Traffic      & 3      & 328    & 27         & 0       & 0    & 0         \\
All          & 17     & 332    & 27        & 4      & 320    & 31         \\
\hline
Keypoint Distribution & \multicolumn{3}{c|}{44-56}    & \multicolumn{3}{c}{27-27-46} \\
\hline
\end{tabular}
\end{table}

The values of the model hyperparameters for the NODE-ESCM are set as follows. $\lambda = 20$, number of iterations $itr = 100$, hidden layer size $h = 100$, learning rate scheduler parameter $\gamma = 0.10$ and step size for the scheduler $s = 100$. $\mathbf{C}_0$ is initialized as $\mathbf{C}_0 = \mathbf{0}$ and all parameters $\theta_0$ are also initialized as $\mathbf{0}$.

\begin{figure*}[h!]
\centering
\begin{subfigure}{.2\textwidth}
    \centering
    \includegraphics[width=0.98\linewidth]{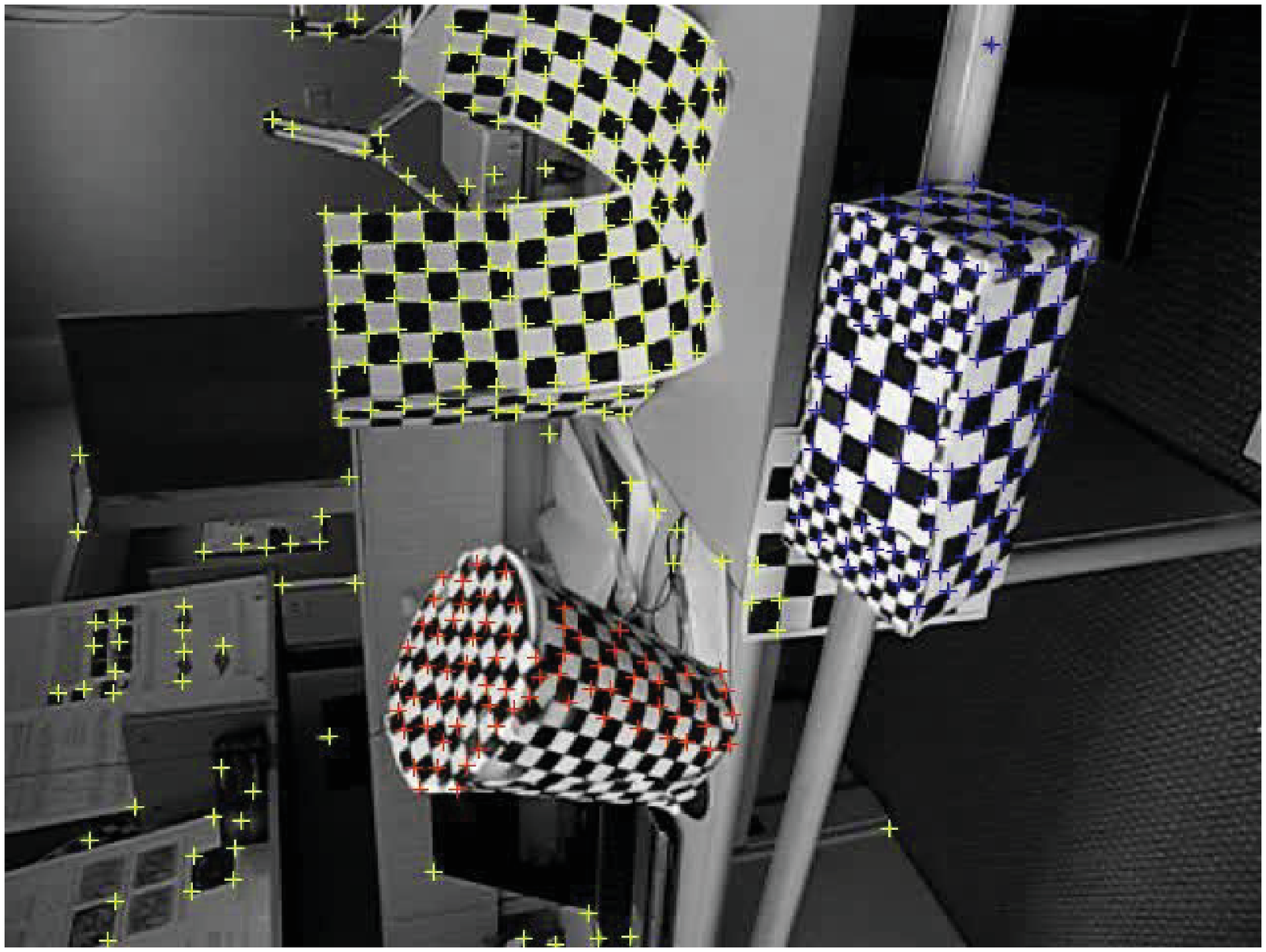}
    \caption{1R2RCTA}
\end{subfigure}
\begin{subfigure}{.2\textwidth}
    \centering
    \includegraphics[width=0.98\linewidth]{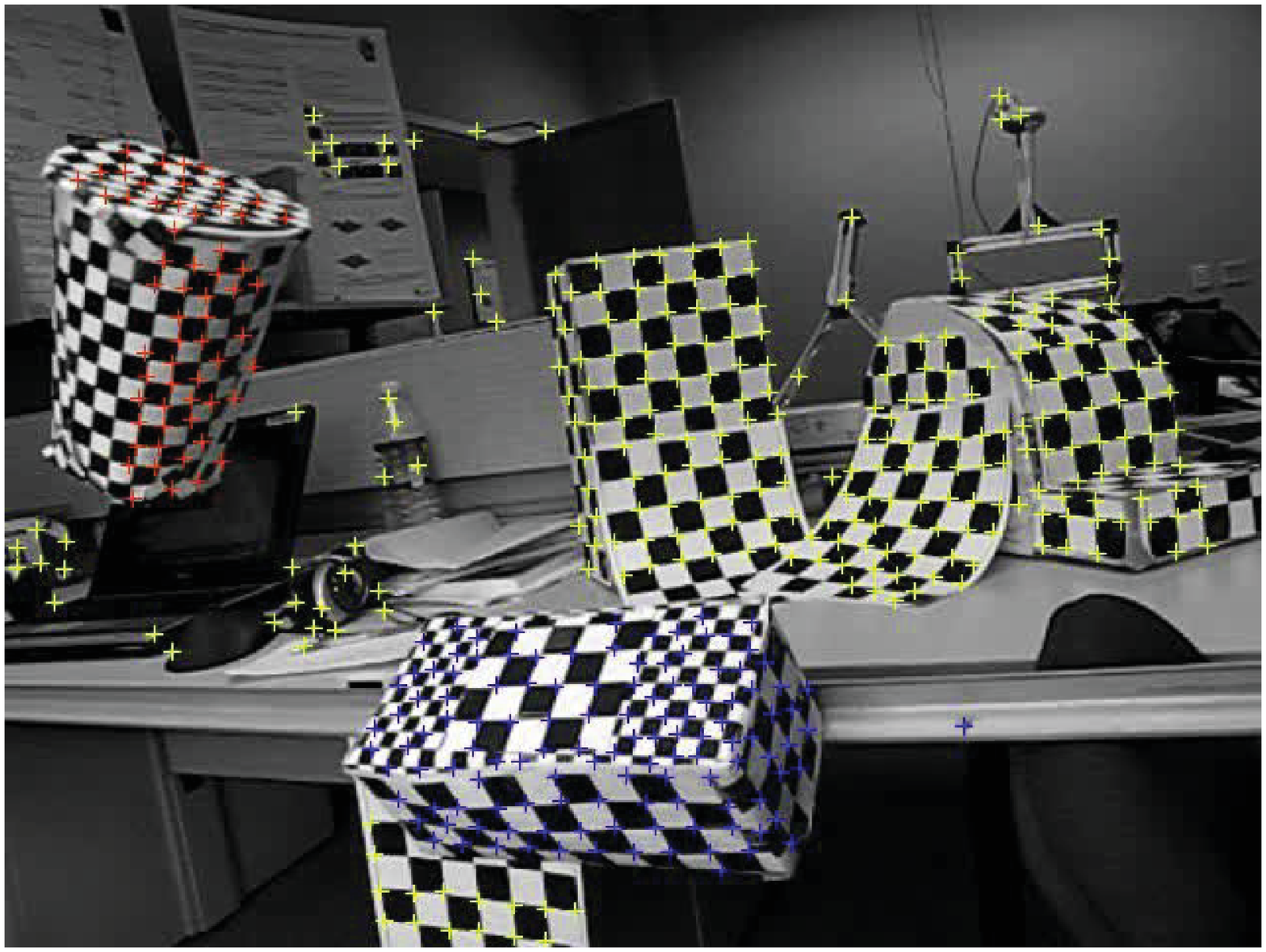}
    \caption{1R2RCTB}
\end{subfigure}
\begin{subfigure}{.2\textwidth}
    \centering
    \includegraphics[width=0.98\linewidth]{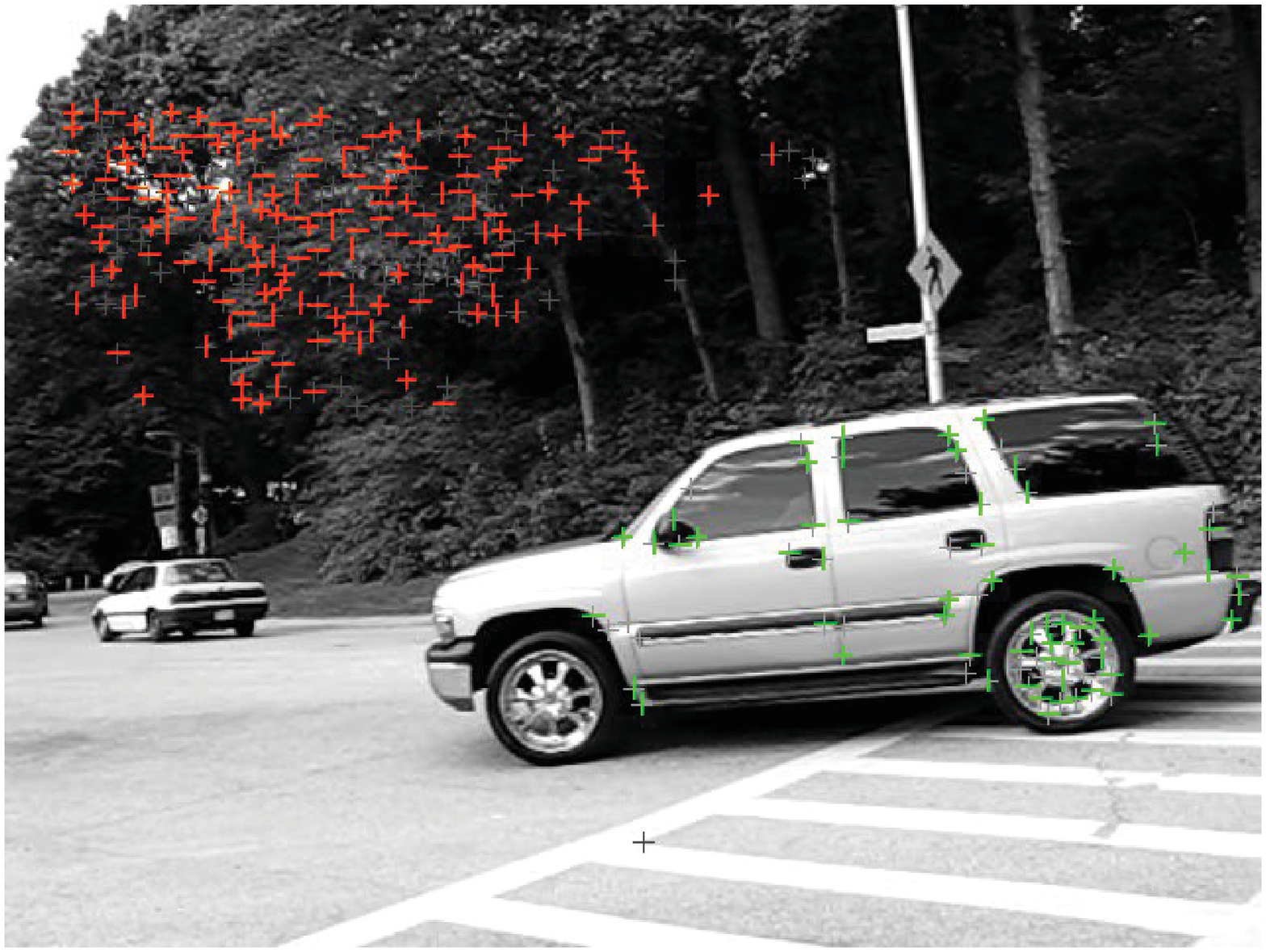}
    \caption{cars1}
\end{subfigure}
\begin{subfigure}{.2\textwidth}
    \centering
    \includegraphics[width=0.98\linewidth]{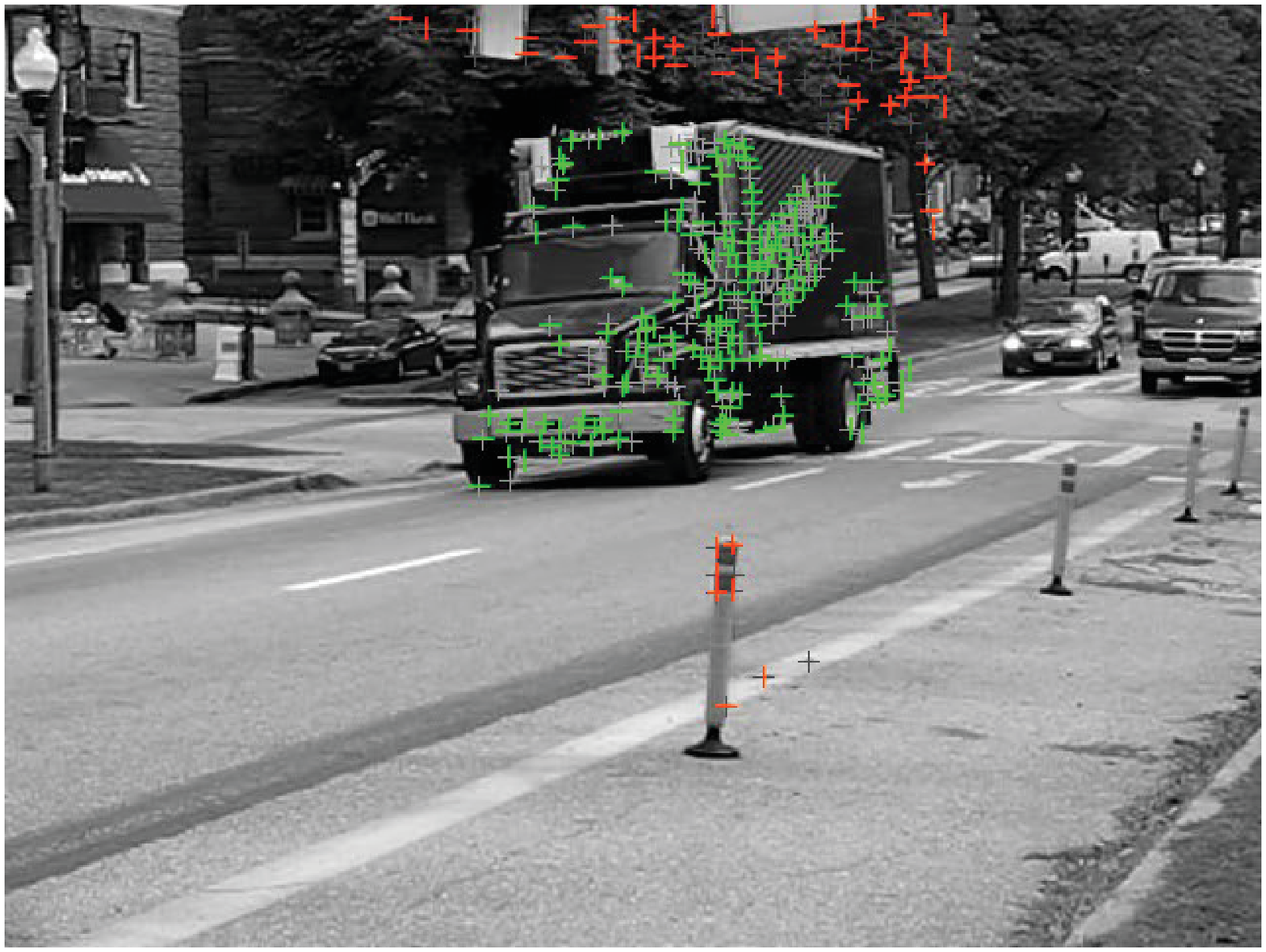}
    \caption{truck1}
\end{subfigure}
\caption{Sample images from some sequences in the
Hopkins 155 database with tracked points superimposed.}\label{fig:motion}
\end{figure*}

\subsubsection{Task}
For the Hopkins 155 dataset, we aim to cluster the $300$ keypoints for each of the $21$ sequences into  $2$ or $3$ low-dimensional subspaces. 
We compute the test error of the NODE-ESCM using the mis-classification rate and make a comparison with other competing subspace clustering methods including the SSC , AFFECT, CESM  and LSTM-ESCM.

\subsubsection{Empirical Results}
The performance of various subspace clustering methods is presented in Table~\ref{Tab:2} and Table~\ref{Tab:3}. For the linear methods, 
we apply 3 compressed sensing approaches: basis pursuit (BP) algorithm \citep{Elhamifar09SparseSC,Elhamifar2013,Chen1998}, orthogonal matching pursuit (OMP) algorithm \citep{Dyer2013,You2016,Pati1993} and accelerated orthogonal least squares (AOLS) algorithm \citep{Hashemi2017,Hashemi2016} with tunable parameter $L=1, 2, 3$. Generally speaking, Table~\ref{Tab:2} reveals that the SCC has a smaller test error than the AFFECT and CESM under the BP and OMP algorithms while the CESM is better than SCC and AFFECT under all three versions of the AOLS algorithm. 
In addition, linear methods based on the OMP algorithm are the most time-efficient. Comparing the linear CESM and nonlinear LSTM-ESCM, we cannot see the advantage of considering the long short-term memory in reducing test error except that the runtime is generally smaller. Among all 5 methods, our NODE-ESCM achieves the smallest test error of $7.86\%$ whereas the test errors of the majority of the competing methods are $3$ to $5$ times bigger, although the runtime of our model is not small. Moreover, the consistency property of the ODE and neural ODE is hence revealed in the NODE-ESCM when learning the representations of the data points in terms of the complicated sequential information, even without irregular time steps.

\begin{table}[tbh]
\centering
\small
\caption{Performance comparisons of the SSC (static subspace clustering), AFFECT and CESM on the Hopkins 155 dataset}
\label{Tab:2}
\begin{tabular}{c|cc|cc|cc}
\hline
\multirow{2}{*}{Learning Method} & \multicolumn{2}{c|}{Static} & \multicolumn{2}{c|}{AFFECT} & \multicolumn{2}{c}{CESM}  \\
                                 & test error (\%) & runtime (s)      & test error (\%) & runtime (s)      & test error (\%) & runtime (s)    \\
\hline
BP                               & 11.07    & 1.59             & 22.40     & 1.39             & 27.86     & 3.11          \\
OMP                              & 25.93    & $\mathbf{0.08}$            & 26.60    & $\mathbf{0.08}$             & 28.07     & 0.09          \\
AOLS ($L=1$)                     & 38.00    & 2.27             & 42.33     & 2.35             & 30.87     & 1.33           \\
AOLS ($L=2$)                     & 23.93     & 3.75              & 31.93     & 1.50            & 23.60     & 1.53           \\
AOLS ($L=3$)                     & 39.27     & 5.27             & 34.93     & 1.82             & 31.47     & 1.55           \\
\hline
\end{tabular}
\end{table}

\begin{table}[tbh]
\centering
\small
\caption{Performance of the LSTM-ESCM and NODE-ESCM on the Hopkins 155 dataset}
\label{Tab:3}
\begin{tabular}{cc|cc}
\hline
\multicolumn{2}{c|}{LSTM-ESCM} & 
\multicolumn{2}{c}{NODE-ESCM}\\
             test error (\%)           & runtime (s)    &test error (\%) & runtime (s)            \\
\hline
31.16    & 0.15                & $\mathbf{7.86}$               & 2.38                \\
\hline
\end{tabular}
\end{table}

\subsection{Empirical Study 2: Ocean Water Mass Clustering}\label{Sec:4:3}
\subsubsection{Experimental Setup}
In this subsection, we investigate the performance of the NODE-ESCM on irregular time steps in an ocean water mass study. We are interested in the ocean temperature and salinity profiles of ocean water mass clustering as water mass is heavily impacted by both temperature and salinity. It is known that water mass contributes to the global climate change, seasonal climatological variations, ocean biogeochemical change and ocean circulation, along with its impact on oxygen and organism transport. The ocean temperature and salinity data are extracted from the ocean water mass study of the Array for Real-time Geostrophic Oceanography (Argo) program, and are made freely available by the International Argo Program and the national programs that contribute to it  (\url{http://www.argo.ucsd.edu}, \url{http://argo.jcommops.org}) \citep{ArgoData}.

The data were collected by the Argo ocean observatory system which consisted of more than $3000$ floats. These floats provided at least  $100,000$ profiles per year and had a 10-day cycle between the ocean surface and $2000$ meters depth. The temperature and salinity measurements were taken in the changing depths between January 2004 and December 2015. All data points were normalized to their monthly averages. In this paper, we are interested in temperature and salinity data from an area near the coast of South Africa, where the Indian Ocean meets the Atlantic Ocean. The latitude of this area is between $25^\circ S$ to $55^\circ S$ and the longitude is between $10^\circ W$ and $60^\circ E$. At the depth level measured at 1000 dbar, a $48$-dimensional feature vector is formed at each gridded location for the normalized temperature and normalized salinity over $24$ months. In total, there are 1684 gridded locations. Now, at each time instance, we construct a matrix $\mathbf{X}_t\in\mathbb{R}^{48\times 1684}$ for one data point. Then, for the evolving sequence, all time steps are chosen to be two years. That is, $t=1$ for Jan 2004 - Dec 2005, $t=2$ for Jan 2006 - Dec 2007, $t=3$ for Jan 2008 - Dec 2009, $t=4$ for Jan 2010 - Dec 2011, $t=5$ for Jan 2012 - Dec 2013 and $t=6$ for Jan 2014 - Dec 2015.

In this study, we use two datasets to assess the performance of our proposed method. In the first dataset, we use the full time series, i.e., $\mathcal{X}_1 = \{\mathbf{X}_t\}_{t=1}^6$ while in the second dataset, 4 time instances are randomly selected without replacement from the 6 time instances and we have $\mathcal{X}_2 = \{\mathbf{X}_{t=1, 3, 4, 6}\}$. Obviously, the time series data in the first dataset are equally-spaced and the time series data in the second dataset are  not. The hyperparameters are set to $\lambda = 20.0$, $h_1 = h_2 = 100$, $\gamma = 0.1$, $\delta = 100$ using the grid search and $itr = 100$. As these two datasets are unlabelled, the data are not ground truth.

\subsubsection{Task}
We shall use our proposed NODE-ESCM to cluster the ocean water mass in the two datasets and evaluate its performance in the cases of both equally-spaced and irregular time steps. We cluster the data into four groups because of no ground truth. These four clusters represent (1) the Agulhas currents, (2) the Antarctic intermediate water (AAIW), (3) the circumpolar deep water mass, and (4) other water masses in the selected area. As labelling is not used  in the datasets, the performance of  our proposed method is assessed by the convergence of training error and the clusters of the temperature and salinity. Whether the NODE-ESCM can capture complicated evolving sequential information can be revealed. Specifically, training errors are computed using the first term of Eq. \eqref{Eq:20} and the first term of Eq. \eqref{Eq:20a} for the first and the second datasets, respectively.

\subsubsection{Empirical Results} 
\begin{figure}[h!]
\begin{subfigure}{.5\textwidth}
    \centering
    \includegraphics[trim={0.6cm, 0.65cm, 0.6cm, 0cm},clip,width=0.98\linewidth]{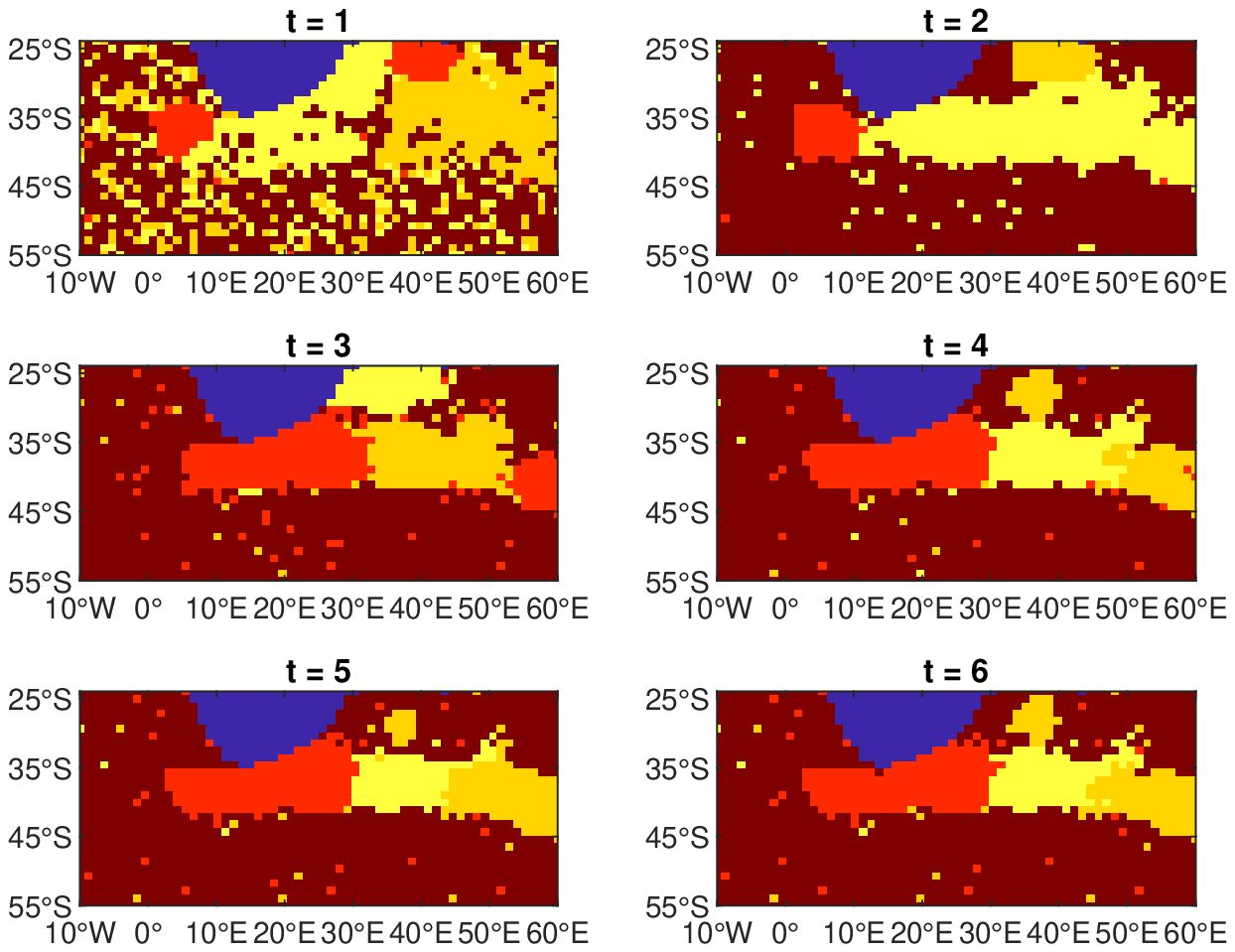}
    \caption{NODE-ESCM: Use full time instances $t = 1, 2, 3, 4, 5, 6$ }
    \label{EvoNeuralODE_full_T_cluster}
\end{subfigure}
\begin{subfigure}{.5\textwidth}
    \centering
    \includegraphics[trim={0.6cm, 0.65cm, 0.6cm, 0cm},clip,width=0.98\linewidth]{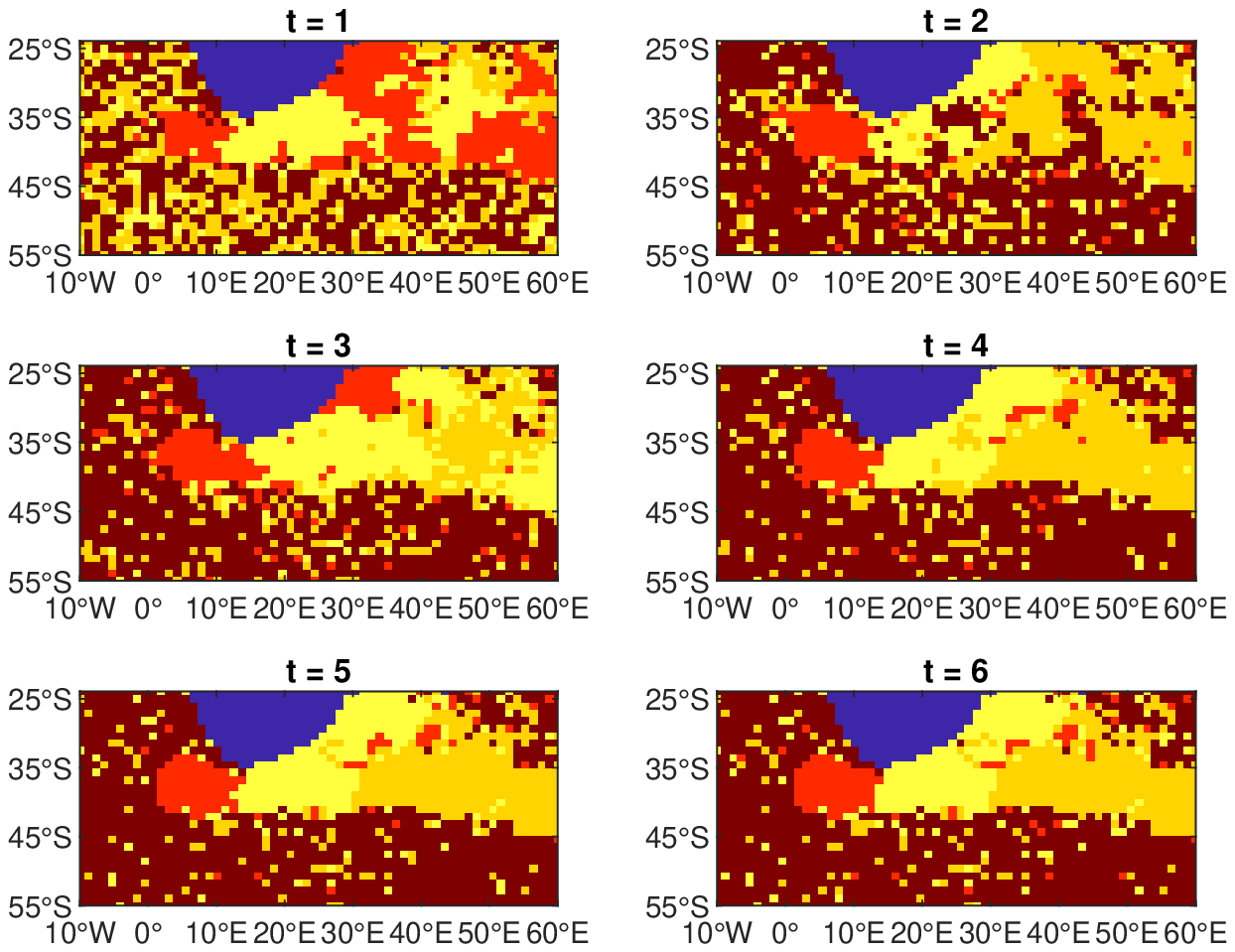}
    \label{EvoNeuralODE_part_T_cluster}
    \caption{NODE-ESCM: Use partial time instances $t = 1, 3, 4, 6$}
    
\end{subfigure}\\
\begin{subfigure}{.5\textwidth}
    \centering
    \includegraphics[width=0.9\linewidth]{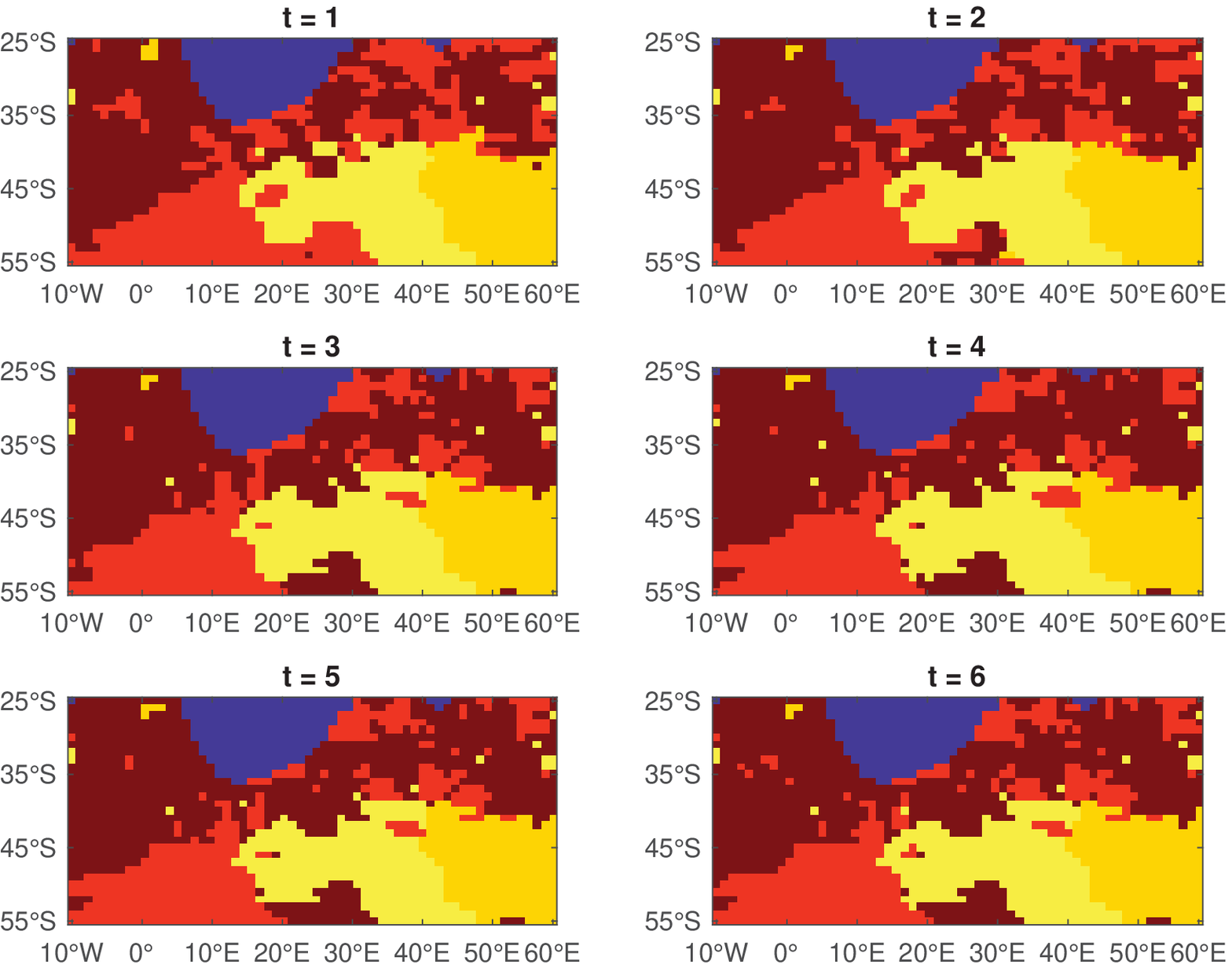}
    \caption{LSTM-ESCM: Use full time instances $t = 1, 2, 3, 4, 5, 6$}
    \label{LSTM-ECSM_full_T_cluster}
\end{subfigure}
\begin{subfigure}{.5\textwidth}
    \centering
    \includegraphics[width=0.9\linewidth]{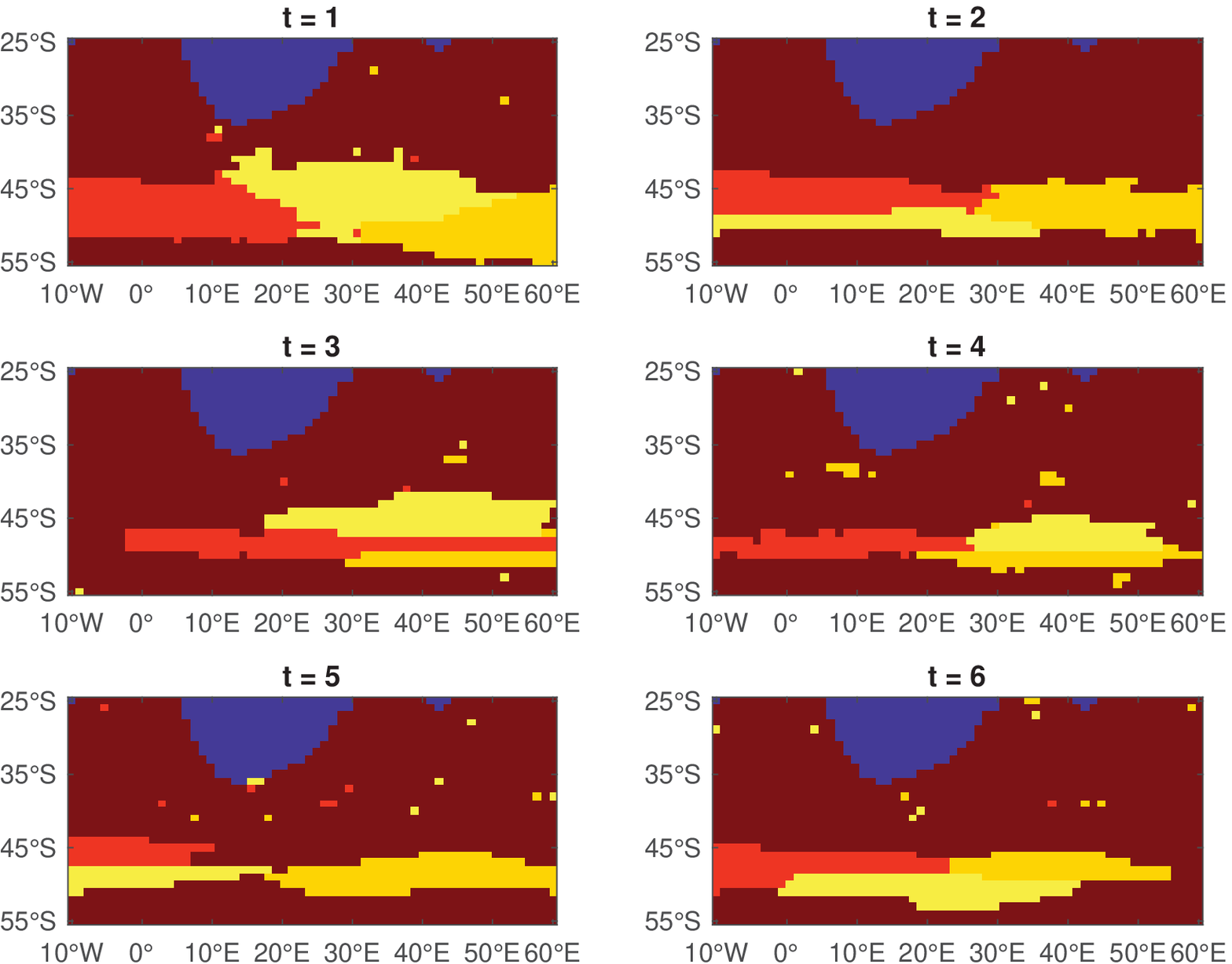}
    \caption{CESM-AOLS-L3: Use full time instances $t = 1, 2, 3, 4, 5, 6$}
    \label{LSTM-ECSM_full_T_cluster}
\end{subfigure}\\
\caption{Clusters of the ocean water by predicted masses: red = Agulhas currents, yellow = AAIW, brown = circumpolar deep water,  orange = other waters and blue = land}\label{Fig:3} 
\end{figure}

Figure~\ref{Fig:3} displays the clusters of ocean water mass at different time steps obtained from different clustering methods. Clusters in red, yellow, brown and orange represent the Agulhas water, AAIW, circumpolar deep water and other waters, respectively. The blue area is land. Obviously, clusters from the NODE-ESCM exhibit a clear evolving sequential relation among all time steps in both datasets, $\mathcal{X}_1$ and $\mathcal{X}_2$. The pattern and shape of the clusters are relatively similar, where the red, yellow and orange clusters distribute from $0^\circ$ to $60^\circ E$ and from $25^\circ S$ to $45^\circ S$, from the west to the east. This finding is consistent with the discoveries in environmental science \citep{Liu2021}.
The NODE-ESCM is able to capture the complicated clustering patterns which are observed from the shapes of the clusters. In 
Figure~\ref{Fig:3}(b),
the clusters of ocean water mass at time steps $t=2$ and $t=5$ are successfully recovered. We can see that there are some similarities in the distribution of clusters to the corresponding images in Figure~\ref{Fig:3}(a).
In fact, during the training process, we treat $t=2,5$ as missing time steps and exclude them from the loss function. This results in some noisy patterns in  Figure~\ref{Fig:3}(b).
Generally speaking, the NODE-ESCM can identify different clusters of ocean water mass. From the locations of these water masses, it is known that the circumpolar deep water mass is generally located beyond $45^\circ S$ and the NODE-ESCM can provide an accurate clustering using $\mathcal{X}_2$. 

As a comparison, clusters obtained using the LSTM-ESCM and CESM with AOLS ($L=3$) (CESM-AOLS-L3) are presented in Figures~\ref{Fig:3}(c) and~\ref{Fig:3}(d). It is obvious that the LSTM-ESCM fails to recognise the circumpolar deep water while the CESM-AOLS-L3 is unable to identify the Agulhas currents. 
In fact, the Limpopo River and Orange River, two major rivers in South Africa, enter the Indian Ocean and Atlantic ocean, respectively.
The estuary for the former is located from $30^\circ E$ to $40^\circ E$ and from $25^\circ S$ to $35^\circ S$ and that of the latter is from $0^\circ$ to $10^\circ E$ and from $30^\circ S$ to $40^\circ S$. Therefore, these two estuaries are expected to be in different clusters from the other areas. From Figure~\ref{Fig:3}, we can see that only the  NODE-ESCM and LSTM-ESCM can successfully distinguish these two estuaries. Moreover, unlike the LSTM-ESCM and CESM-AOLS-L3 which require to use the equally-spaced time series data $\mathcal{X}_1$ in the learning procedure, NODE-ESCM can effectively obtain the ocean water mass distribution and provide reasonable explanations using irregularly-sampled time series, $\mathcal{X}_2$. 

In terms of training error convergence, our NODE-ESCM succeeds in the case of time series modelling with irregular time steps. Table~\ref{tab:oceanconv} presents an empirical convergence analysis. For both datasets, $\mathcal{X}_1$ and $\mathcal{X}_2$, convergence is achieved at the $20^{\rm th}$ epoch. The minimum training errors are 0.005130 and 0.004581 for $\mathcal{X}_1$ and $\mathcal{X}_2$, respectively. 

\begin{table}[!h]
    \centering
    \small
    \caption{Convergence of training errors: the performance of NODE-ESCM on full time instances $\mathcal{X}_1 = \{\mathbf{X}_t\}_{t=1}^6$ and partial time instances $\mathcal{X}_2 = \{\mathbf{X}_{t=1, 3, 4, 6}\}$}
    \begin{tabular}{l||c|c}
    \hline
        & Full time instances $\mathcal{X}_1$ & Partial time instances $\mathcal{X}_2$ \\
         \hline \hline
         Converged training error& 0.005130 & 0.004581 \\
         \hline
         Convergence epoch& 20 & 20 \\
         \hline
    \end{tabular}
    \label{tab:oceanconv}
\end{table}

To conclude, the performance is satisfactory and the results are valid when the NODE-ESCM is used to time series data collected at irregular time steps. This method is even more capable of processing complicated non-linear information when the assumption on the equally-spaced time steps is violated. Therefore, our proposed NODE-ESCM can conduct data analysis tasks with more flexibility on the time domain when all other evolutionary subspace clustering methods are not able to. When given information is limited, the NODE-ESCM may still be able to discover the general pattern of the data.

\subsection{Empirical Study 3: Female's Welfare Performance Clustering in G20 Countries}
\subsubsection{Experimental Setup}
Female's welfare and well-being are classical research topics in social science \citep{Klumb2004,GATTA2008}. They have an important perspective on gender study. In recent global and regional social movements, such as ``Me Too'', it brings to our attention that the economic and political power of a nation may not have a positive correlation with its female's welfare. In this study, we investigate the female's welfare performance in 19 G20 countries and the European Union (EU) is excluded from the study. These countries are Argentina, Australia, Brazil, Canada, China, France, Germany, India, Indonesia, Italy, Japan, Mexico, Russia, Saudi Arabia, South Africa, South Korea, Turkey, the United Kingdom and the United States.

The data we use for clustering purposes are obtained from the World Health Organization (WHO)\footnote{\url{https://www.who.int/data/collections}} and the World Bank\footnote{\url{https://data.worldbank.org/indicator}}. They are yearly data of 15 indicators collected by these two organizations between 2006 and 2016 and the indicators are described in Table~\ref{ind}.
\begin{table}[!h]
    \centering
    \small
    \caption{Female welfare indicators}
    \begin{tabular}{|c|p{0.7\textwidth}|p{0.2\textwidth}|}
    \hline 
        \textbf{No.} & \textbf{Indicator} & \textbf{Source} \\
         \hline
          1 & Male to female life expectancy at birth ratio ($\%$) & WHO \\  \hline 
          2 & 15+ year-old male to female employment ratio ($\%$) & The World Bank \\ \hline 
          3 & Male to female vulnerable employment ratio ($\%$) & The World Bank \\ \hline
          4 & 0-14 year-old male to female population ratio ($\%$) & The World Bank \\ \hline  
          5 & Primary school enrollment (gross), gender parity index (GPI) & The World Bank \\ \hline
          6 & Adolescent fertility rate (births per 1,000 women ages 15-19) & The World Bank \\ \hline
          7 & Proportion of seats held by women in national parliaments ($\%$) & The World Bank \\ \hline
          8 & Male to female unemployment ratio ($\%$) & The World Bank \\ \hline
          9 & Maternal mortality ratio (modeled estimate, per 100,000 live births) & The World Bank \\ \hline
          10 & Male to female contributing family workers ratio ($\%$) & The World Bank \\ \hline
          11 & Male to female salaried and waged worker ratio ($\%$) & The World Bank \\ \hline
          12 & Male to female employment in agriculture ratio ($\%$) & The World Bank \\ \hline
          13 & Male to female employment in industry ratio ($\%$) & The World Bank \\ \hline
          14 & Male to female employment in service ratio ($\%$) & The World Bank \\ \hline
          15 & Non-pregnant female anaemia rate ($\%$) & WHO \\ \hline   \end{tabular}
    \label{ind}
\end{table}
The reason for selecting these 15 indicators is that female's welfare is a  complicated and highly debatable topic and we wish to use as many aspects and data as possible in our study.  
The aspects include health (Indicators $1, 6, 9, 15$), education (Indicator $5$), economy (Indicators $2, 3, 8, 10, 11, 12, 13, 14$) and politics  (Indicator $7$). Indicators 1-14 are related to gender equality and only Indicator 15 emphasizes on female's health.

Data are available for all 19 G20 countries and hence the dataset contains no missing values. This dataset is $\mathcal{X} = \{\mathbf{X}_t \in \mathbb{R}^{15\times 19}\}$ with $t=1$ for Year 2006, $t=2$ for Year 2007, and so on. For each time period, there are 15 features and 19 countries. In this study, the hyperparameters in the NODE-ESCM are set to $\lambda = 0.01$, $h_1 = h_2 = 100$, $\gamma = 0.1$, $\delta = 100$ using the grid search and $itr = 100$. Again, the data are not ground truth as this dataset is unlabelled.

\subsubsection{Task}
We aim to analyze the data and cluster the 19 G20 countries into 3 distinct groups according to the similarity of their female's welfare situations. The clustering is performed using our proposed NODE-ESCM. The performance of the method is evaluated by the general consistency of our results with the common knowledge in social science, and the new patterns that our method can identify.

\subsubsection{Empirical Results}
\begin{figure*}[t!p]
\centering
\begin{subfigure}[b]{.4\textwidth} 
    \centering
    \captionsetup{font={tt, bf, small}}
    \caption*{2006 ($t$ = 1)}
    \includegraphics[width=0.95\linewidth]{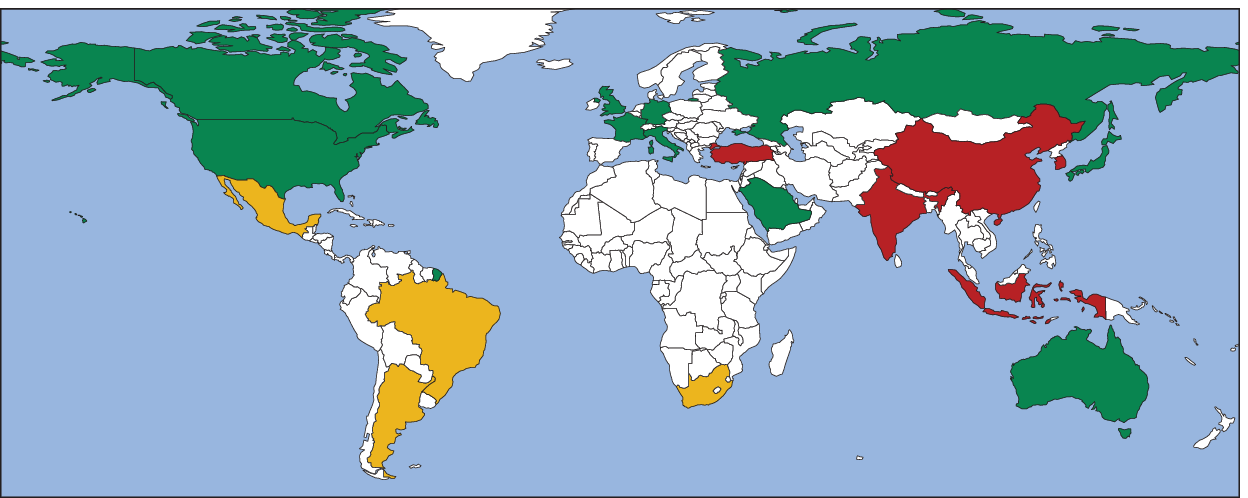}
    \label{2006}
\end{subfigure}
\begin{subfigure}[b]{.4\textwidth}
    \centering
    \captionsetup{font={tt, bf, small}}
    \caption*{2007 ($t$ = 2)}
    \includegraphics[width=0.95\linewidth]{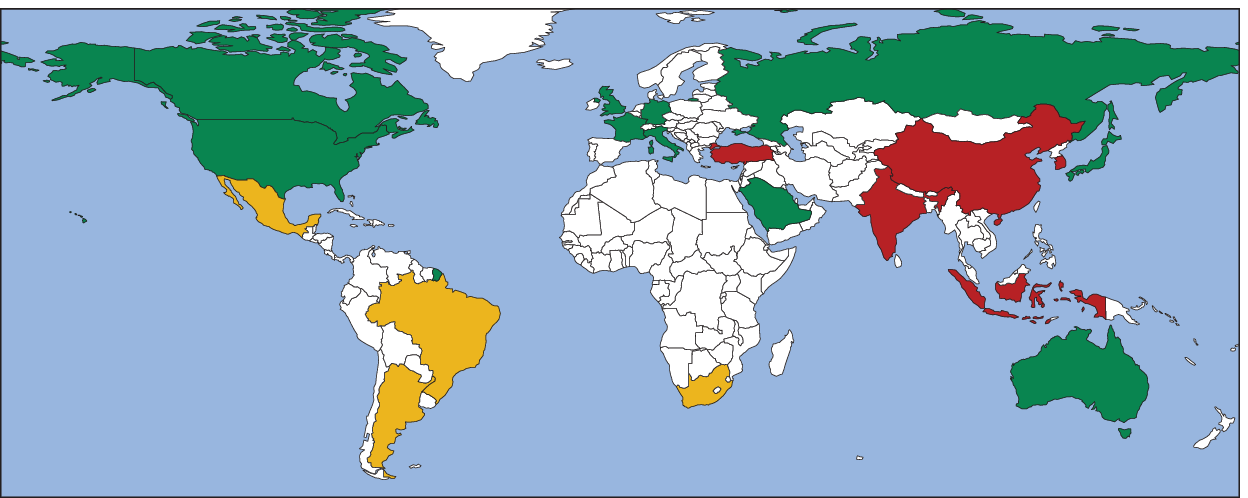}
    \label{2007}
\end{subfigure} \\[8pt]
\begin{subfigure}[b]{.4\textwidth}
    \centering
    \captionsetup{font={tt, bf, small}}
    \caption*{2008 ($t$ = 3)}
    \includegraphics[width=0.95\linewidth]{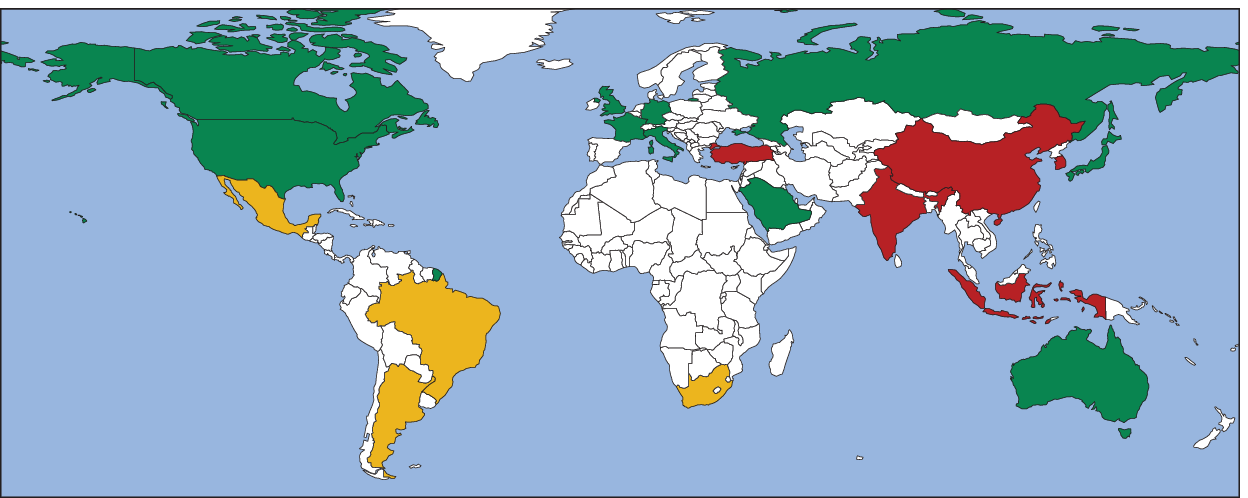}
    \label{2008}
\end{subfigure} 
\begin{subfigure}[b]{.4\textwidth}
    \centering
    \captionsetup{font={tt, bf, small}}
    \caption*{2009 ($t$ = 4)}
    \includegraphics[width=0.95\linewidth]{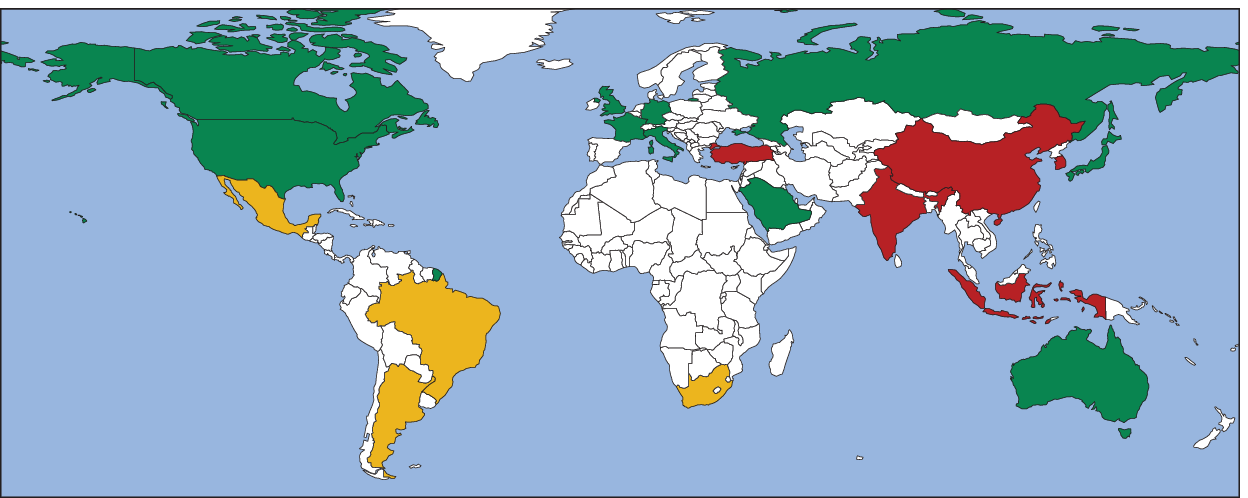}
    \label{2009}
\end{subfigure} \\[8pt]
\begin{subfigure}[b]{.4\textwidth}
    \centering
    \captionsetup{font={tt, bf, small}}
    \caption*{2010 ($t$ = 5)}
    \includegraphics[width=0.95\linewidth]{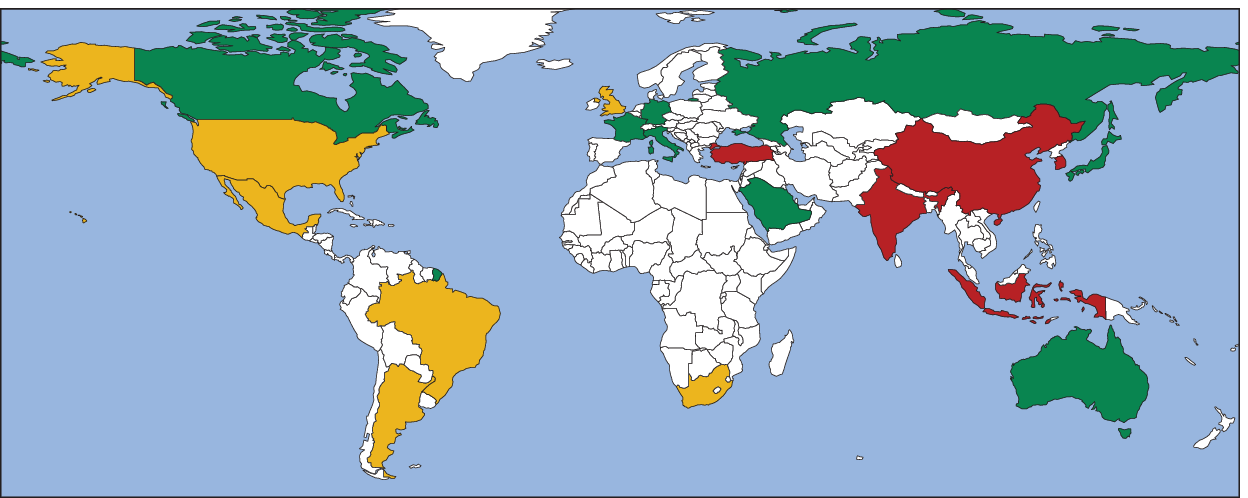}
    \label{2010}
\end{subfigure} 
\begin{subfigure}[b]{.4\textwidth}
    \centering
    \captionsetup{font={tt, bf, small}}
    \caption*{2011 ($t$ = 6)}
    \includegraphics[width=0.95\linewidth]{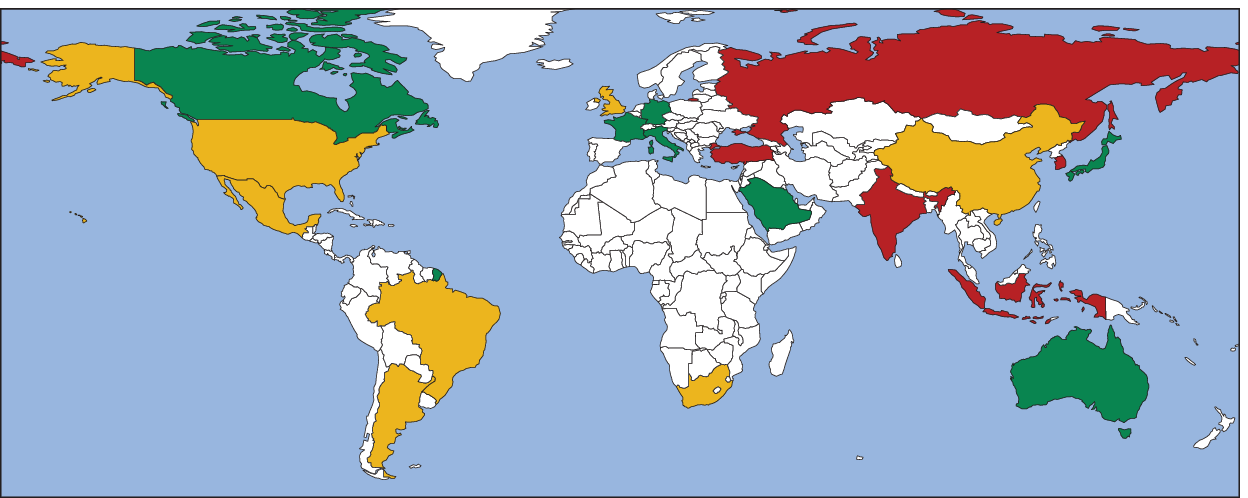}
    \label{2011}
\end{subfigure} \\[8pt]
\begin{subfigure}[b]{.4\textwidth}
    \centering
    \captionsetup{font={tt, bf, small}}
    \caption*{2012 ($t$ = 7)}
    \includegraphics[width=0.95\linewidth]{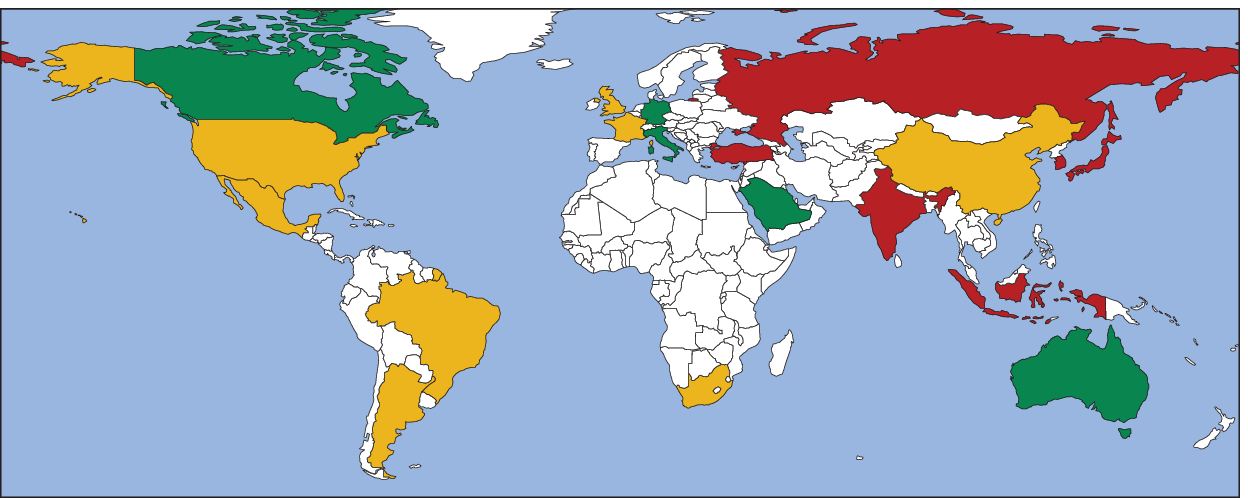}
    \label{2012}
\end{subfigure}
\begin{subfigure}[b]{.4\textwidth}
    \centering
    \captionsetup{font={tt, bf, small}}
    \caption*{2013 ($t$ = 8)}
    \includegraphics[width=0.95\linewidth]{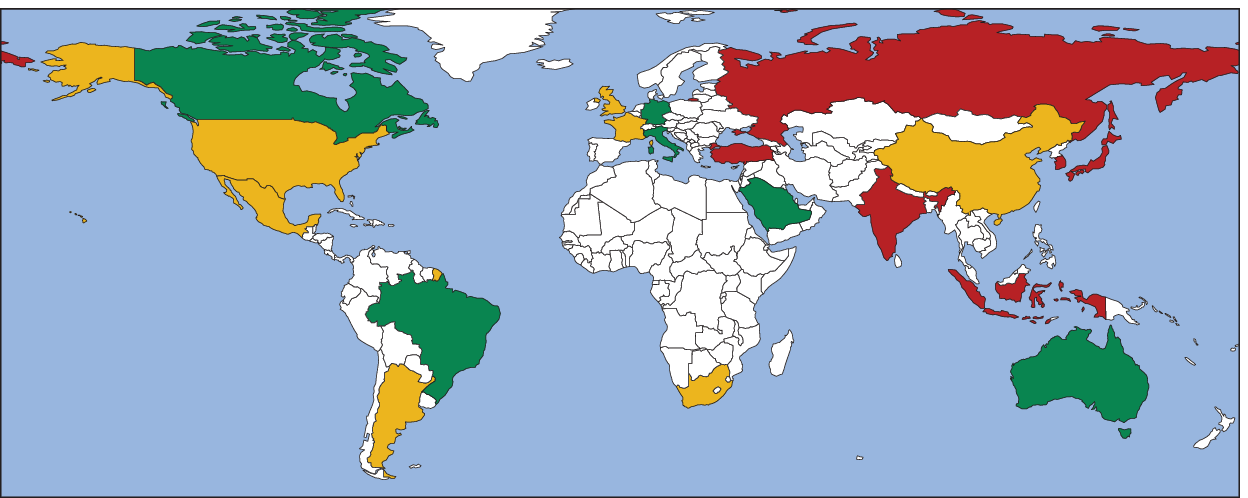}
    \label{2013}
\end{subfigure} \\[8pt]
\begin{subfigure}[b]{.4\textwidth}
    \centering
    \captionsetup{font={tt, bf, small}}
    \caption*{2014 ($t$ = 9)}
    \includegraphics[width=0.95\linewidth]{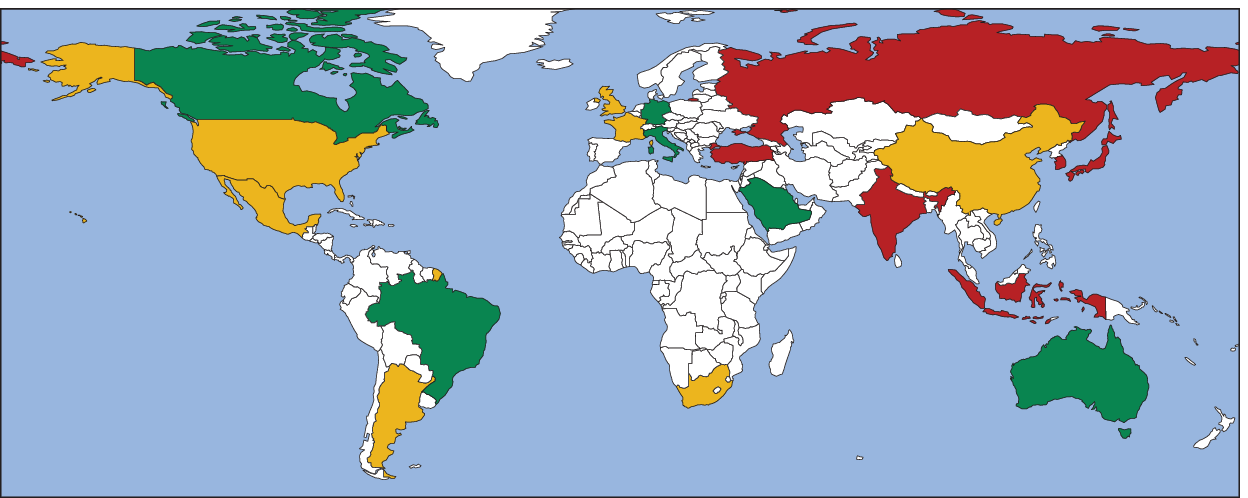}
    \label{2014}
\end{subfigure} 
\begin{subfigure}[b]{.4\textwidth}
    \centering
    \captionsetup{font={tt, bf, small}}
    \caption*{2015 ($t$ = 10)}
    \includegraphics[width=0.95\linewidth]{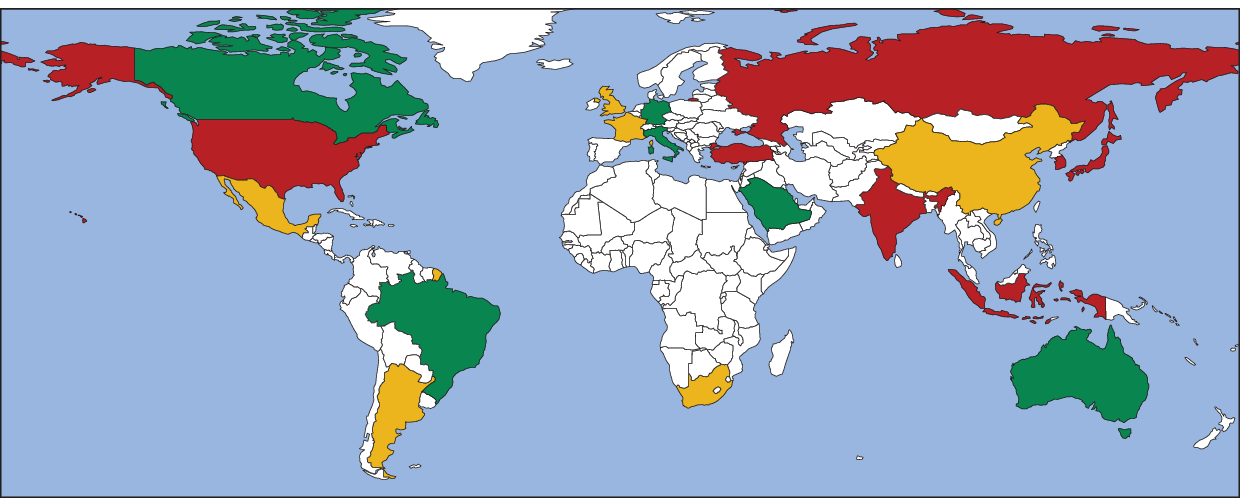}
    \label{2015}
\end{subfigure} \\[8pt]
\begin{subfigure}[b]{.4\textwidth}
    \centering
    \captionsetup{font={tt, bf, small}}
    \caption*{2016 ($t$ = 11)}
    \includegraphics[width=0.95\linewidth]{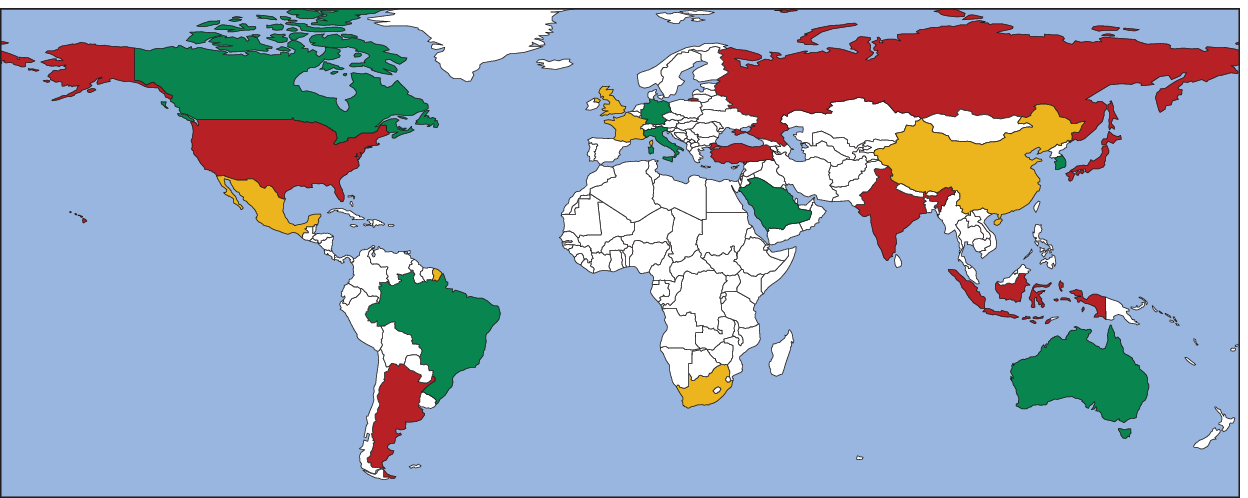}
    \label{2016}
\end{subfigure} 
\caption{Female's welfare performance clustering for 19 G20 countries from 2006 to 2016.}\label{fig:gender}
\end{figure*}

Based on the NODE-ESCM, the 19 G20 countries are clustered into three groups on a yearly basis. In Figure~\ref{fig:gender}, clusters in green, yellow and red represent the good, medium and poor performance groups, respectively. 
Generally speaking, the changes in female's welfare performance of a country over the study period were slow. This observation is consistent with the common sense in social science that social changes are always difficult and very slow. In this study, our method identifies 10 countries whose female's welfare performance did not change over the entire study period. These countries are Australia, Canada, Germany, India, Indonesia, Italy, Mexico, Saudi Arabia, South Africa and Turkey. The countries that showed changes are Argentina, Brazil, China, France, Japan, Russia, South Korea, the UK and the US.

Between 2006 ($t=1$) and 2009 $(t=4)$, there was no change in female's welfare performance in all 19 countries. Australia, Canada, France, Germany, Italy, Japan, Russia, Saudi Arabia, the UK and the US, were in the green cluster, Argentina,  Brazil, Mexico and South Africa were in the yellow cluster, and China, India, Indonesia, South Korea and Turkey were in the red cluster. Interestingly, the clusters of the countries generally reveal the cultural groups. For example, countries with Asian cultures and countries historically heavily impacted by the Asian cultures were likely in the same cluster. Likewise, Latin American countries were also in the same cluster and countries vastly rooted in the European cultures also group together. Having experienced in westernization socially on female's welfare, Japan, with strong Asian cultures, and Saudi Arabia, as an Islamic country, were in the same cluster as countries with European cultures. South Africa, as a multilingual and ethnically diverse country with a majority of African population, was in the same group with the Latin American countries.

Between 2009 and 2016, the changes in female's welfare performance in some countries were noticeable. In particular, the US changed from the green cluster to the yellow cluster in 2010 and then to the red cluster in 2015 and it is also the only G20 country that underwent two changes in 11 years. In 2016, the US was in the red cluster with Argentina, India, Indonesia, Japan, Russia and Turkey. Hence, the female's welfare performance in the US is alarming. France and the UK moved from the green cluster to the yellow cluster in 2010 and 2012, respectively, while Russia and Japan shifted from the green cluster to the red cluster in 2011 and 2012, respectively. Argentina left the yellow cluster and joined the red cluster in 2016. On the contrary, some countries have improved their female's welfare performance. China entered the yellow cluster from the red cluster in 2011, Brazil moved from the yellow cluster to the green cluster in 2013 and South Korea jumped from the red cluster to the green cluster in 2016. 
The changes in these 9 countries can be interpreted as the consequences of legal and political progress on gender equality and female protection. Taking the US as an example, the Lily Ledbetter Fair Pay Act introduced in 2009 allows victims, usually women, to file pay discrimination complaints with the government against their employers. The ``Me Too'' movement founded by Tarana Burke in 2006 has raised the awareness of women who had been abused, and this impact of the movement continues to increase with the popularity of social media. Though South Korea did not show a significant change in its female's welfare performance before 2016, the election of the first female president Park Geun-hye in 2012 was a big advance in female's welfare. Eventually, South Korea joined Australia, Brazil, Canada, Germany, Italy and Saudi Arabia as a member of the green cluster in 2016. 

In this study, our proposed NODE-ESCM successfully captures the sequential information on female's welfare performance in 19 G20 countries over 11 years. Our findings agree with the expectations from the field of social science. In addition, this method can reveal the dynamic of female's welfare performance from 2010 to 2016 for Argentina, Brazil, China, France, 
Japan, Russia, South Korea, the UK and the US. An interesting relationship between the social changes and cultures is also unveiled.

\section{Conclusion}\label{Sec:5}
In this paper, we propose the neural ODE evolutionary subspace clustering method (NODE-ESCM) to time series data. Unlike many existing clustering methods, NODE-ESCM does not assume the equally-spaced time steps. Moreover, its affinity matrix is the solution of a well-defined NODE, which can be found without using any explicit algorithms. Given an initial value and the learned first derivative, which is a neural network, the affinity matrix can be obtained at any time point. The well-defined NODE also possesses a nice theoretical property that is consistency. In the learning procedure, we use the self-expressive constraint in the loss function. Experimental results demonstrate the effectiveness and superiority of this  method. NODE-ESCM outperforms all baseline methods including the state-of-the-art ESCMs in our simulation and empirical studies. In the simulation study, the NODE-ESCM achieves a higher accuracy than the nonlinear LSTM-ESCM and is robust to the hyperparameters. In the empirical study, the NODE-ESCM has the smallest test error than the SCC, AFFECT, CESM and LSTM-ESCM in the study of real-time motion segmentation data. In the ocean water mass analysis, the NODE-ESCM can successfully capture the clustering features of the ocean water mass around a South Africa coast where the Atlantic Ocean and Indian Ocean meet, while the CESM and LSTM-ESCM cannot. In addition, in the case of irregular time steps where data in two randomly selected time points are treated as missing, the NODE-ESCM is able to provide a reasonable explanation of the ocean water mass distribution. In the female's welfare study, the NODE-ESCM clusters the 19 G20 countries into 3 groups according to their performance on female's welfare between 2006 and 2016. It can generally rank the performance of the 19 countries in these three clusters from good to poor. Our findings are consistent with the common knowledge in social science and news stories, and reasonably explain the changes in female's welfare performance of some countries during the 11 years study period. In addition, we also discover that the female's welfare of a nation is not positively correlated with its economic and political power. This is also the first gender study using subspace clustering with deep networks. For future work, we will explore more structural information in the data such as graphs. Furthermore, the NODE-ESCM does not consider the volatility of the evolving information in time series. As the NODE assumes the time series to be generally smooth on the time domain, we may use stochastic methods to model the volatility in the affinity matrix in our future work.

\bibliography{reference}

\end{document}